\documentclass[10pt,twocolumn,letterpaper]{article}

\usepackage[pagenumbers]{cvpr} % To force page numbers, e.g. for an arXiv version
\usepackage{algpseudocode}
\usepackage{algorithm}
\usepackage{caption}
\usepackage{cuted}
\usepackage{soul}
\usepackage{multirow}

\usepackage[dvipsnames]{xcolor}

\newcommand{\DDIM}{\textrm{DDIM}}
\newcommand{\LDM}{\textit{LDM}}
\newcommand{\VAE}{\textit{VAE}}
\newcommand{\za}[1]{z^A_{#1}}
\newcommand{\zb}[1]{z^B_{#1}}
\newcommand{\zstar}[1]{z^*_{#1}}
\newcommand{\zstarpred}[1]{\hat{z}^*_{#1}}
\newcommand{\appzb}[1]{\tilde{z}^B_{#1}}
\newcommand{\predzb}[1]{\hat{z}^B_{#1}}

\newcommand{\xa}[1]{x^A_{#1}}

\newcommand{\appxb}[1]{\tilde{x}^B_{#1}}
\newcommand{\pred}[2]{\hat{#1}_{#2}}

\newcommand{\LossSTs}{\mathcal{L}^\textrm{ST}_\textit{str}}
\newcommand{\LossSTa}{\mathcal{L}^\textrm{ST}_\textit{app}}
\newcommand{\LossTOs}{\mathcal{L}^\textrm{TO}_\textit{str}}
\newcommand{\LossTOa}{\mathcal{L}^\textrm{TO}_\textit{app}}
\newcommand{\Loss}{\mathcal{L}}

\definecolor{cvprblue}{rgb}{0.21,0.49,0.74}
\usepackage[pagebackref,breaklinks,colorlinks,citecolor=cvprblue]{hyperref}

\def\paperID{11132} % *** Enter the Paper ID here
\def\confName{CVPR}
\def\confYear{2024}

\title{S2ST: Image-to-Image Translation in the Seed Space of Latent Diffusion}

\author{Or Greenberg\\
GM R\&D\\
{\tt\small or.greenberg@gm.com}
\and
Eran Kishon\\
GM R\&D\\
{\tt\small eran.kishon@gm.com}
\and
Dani Lischinski\\
The Hebrew University\\
{\tt\small danix@mail.huji.ac.il}
}

\begin{document}
\maketitle

\begin{abstract}
Image-to-image translation (I2IT) refers to the process of transforming images from a source domain to a target domain while maintaining a fundamental connection in terms of image content. In the past few years, remarkable advancements in I2IT were achieved by Generative Adversarial Networks (GANs), which nevertheless struggle with translations requiring high precision. Recently, Diffusion Models have established themselves as the engine of choice for image generation.  
In this paper we introduce S2ST, a novel framework designed to accomplish global I2IT in complex photorealistic images, such as day-to-night or clear-to-rain translations of automotive scenes. S2ST operates within the seed space of a Latent Diffusion Model, thereby leveraging the powerful image priors learned by the latter. We show that S2ST surpasses state-of-the-art GAN-based I2IT methods, as well as diffusion-based approaches, for complex automotive scenes, improving fidelity while respecting the target domain's appearance across a variety of domains. Notably, S2ST obviates the necessity for training domain-specific translation networks.
\end{abstract}
    
\section{Introduction}
\label{sec:intro}

%\begin{strip}
\begin{figure}
  \centering
  \includegraphics[width=\columnwidth]{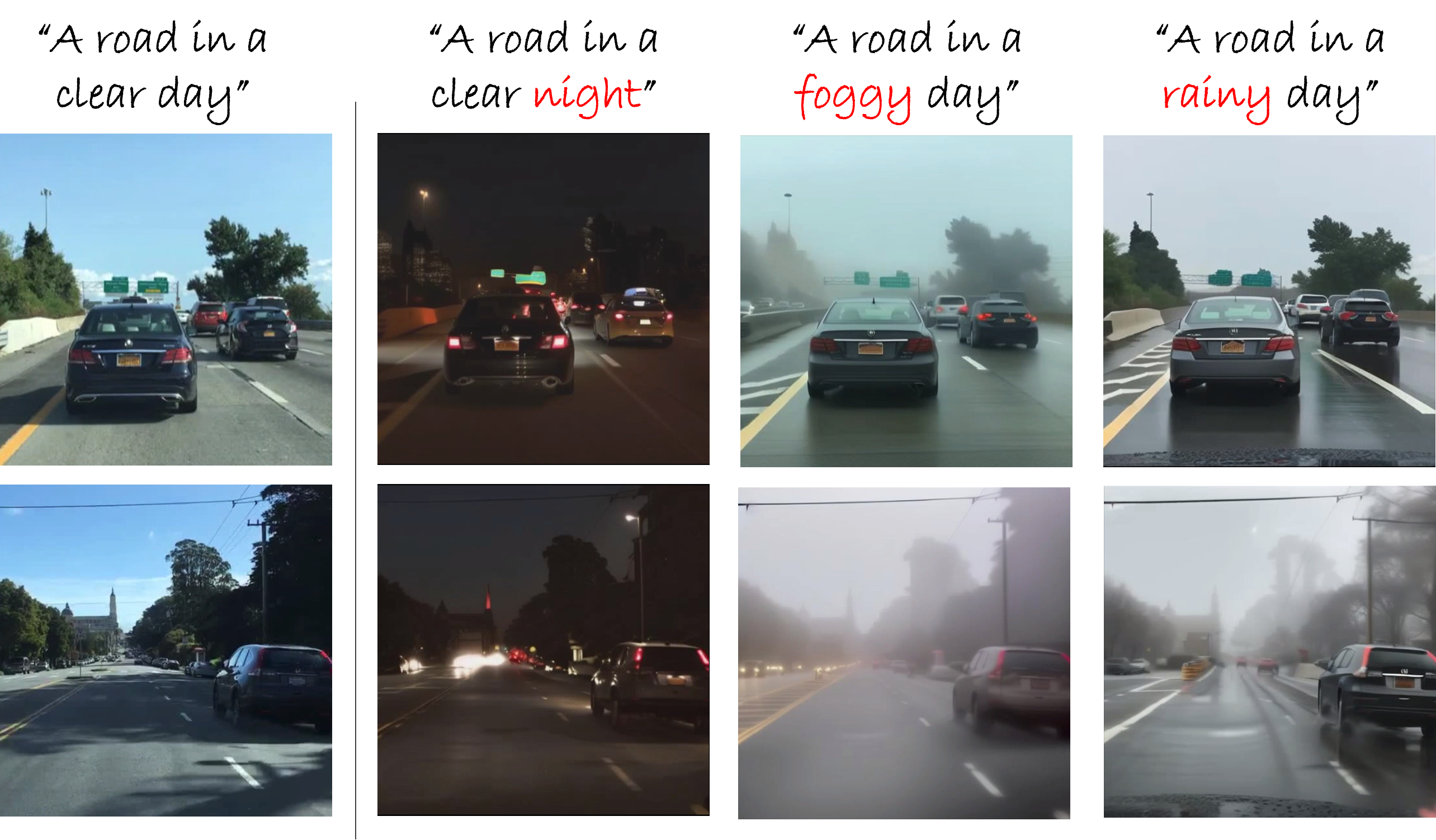}
  \captionof{figure}{\textbf{S2ST: Seed-to-Seed Translation}. We propose a mechanism for multi-domain image-to-image translation of complex (e.g., automotive) scenes performed within the seed space of a Latent Diffusion Model. Here we show real input images with their corresponding translated outputs in different target domains.}
  \label{fig:teaser}
\end{figure}
%\end{strip}    

Image-to-Image Translation (I2IT) refers to the process of transforming an image from one domain to another, while preserving the underlying structure and semantic content of the original image.
Of particular interest is the unpaired I2IT setting, which tackles the translation task without requiring pairs of source and target images for training and validation. In recent years, immense progress in unpaired I2IT has been achieved through the utilization of Generative Adversarial Networks (GANs) \cite{zhu2017unpaired, liu2017unsupervised,  huang2018multimodal, jiang2020tsit, dutta2022seeing}. Despite their impressive results, GAN-based I2IT models face challenges in translation of complex photo-realistic images with high content fidelity requirements, such as day-to-night translation of automotive scenes.
In addition, GANs are typically constrained to learning one-to-one mappings between predefined domains, limiting their flexibility and applicability when translating between multiple domains.

In recent years, text-to-image diffusion models like DALL-E 2~\cite{ramesh2022hierarchical}, Imagen \cite{saharia2022photorealistic}, and Stable Diffusion \cite{rombach2022high}, have demonstrated impressive capabilities in text-guided generation of photo-realistic and diverse synthetic images from random noise. In this paper we use the term \emph{seed space} to denote the space of (potentially latent) noise samples (\emph{seeds}), from which the diffusion sampling process is initiated \cite{samuel2023norm}. In previous research, generative capabilities of diffusion models have been used for advanced image manipulations. While demonstrating remarkable effectiveness in a variety of image manipulations, models designed for global editing often alter the image in a manner that is unpredictable and challenging to control. 

To address the aforementioned challenges, we propose Seed-to-Seed Translation (S2ST), a diffusion-based unpaired I2IT solution that harnesses the powerful priors of a trained diffusion model to perform global edits, focusing on day-to-night and clear-to-adverse weather translations of complex automotive scenes. A few examples are shown in \Cref{fig:teaser}.
High quality unpaired I2IT is crucial for driving simulators and efficient image augmentation during the training of perception algorithms for automotive applications. We show that S2ST outperforms state-of-the-art GAN-based methods in terms of its capacity to learn multiple target domain appearances, while retaining content and structure and eliminating the need for training separate domain-specific networks. 

\begin{figure}[t]
  \centering
  %\fbox{\rule{0pt}{2in} \rule{0.9\linewidth}{0pt}}
   \includegraphics[width=0.9\linewidth]{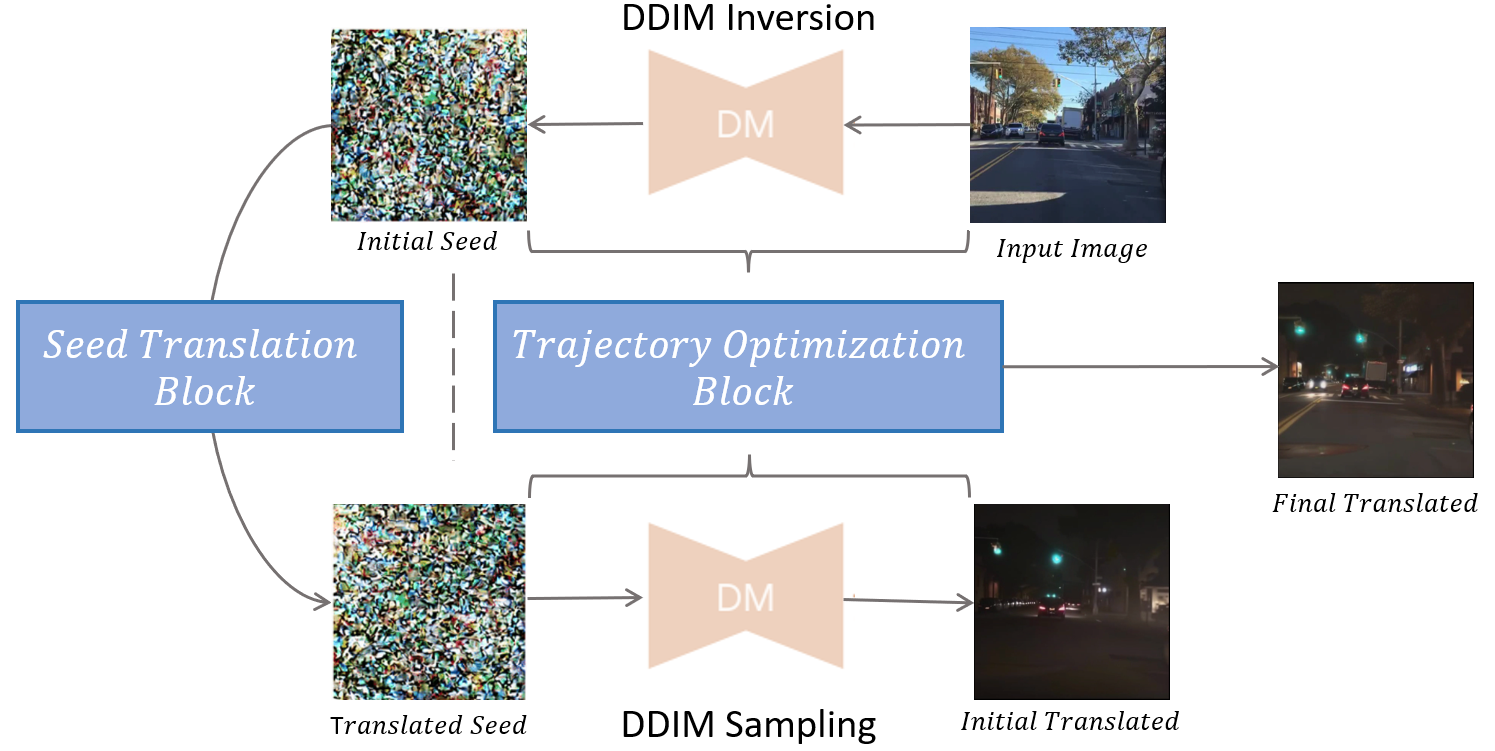}

   \caption{\textbf{S2ST framework overview:} The input image is first inverted to the seed space, where it is translated to the target domain. Next, Trajectory Optimization is carried out to enhance structural similarity between the source image and the translated one.}
   \label{fig:overall}
\end{figure}

To generate photo-realistic target domain images, while adhering to the semantic content and structure of the source input image, we propose a novel 2-step process, depicted in \Cref{fig:overall}, with Stable Diffusion \cite{rombach2022high} as the generative engine: Given a source image, we apply DDIM inversion~\cite{song2020denoising} to obtain an initial seed; next, (1) Seed optimization is applied to the initial seed to obtain a translated seed, aligned with the target domain, while enforcing structural similarity between the source and the generated image; (2) the samples generated by the DDIM sampling process from the translated seed to the ``clean'' latent domain sample (generation trajectory) are optimized using the source inversion trajectory (samples generated from the source by DDIM inversion), to further enhance the structural similarity between the source and translated image.

With a textual description of the target domain, and a few ($\sim$ 5) example images from this domain, our approach can generate a synthetic image that is well-aligned with the target domain, while preserving the structure and content of the input source image. Using both quantitative and qualitative comparisons, we show that our method outperforms SoTA GAN-based I2IT, as well as recent diffusion-based image manipulation methods in terms of domain appearance, photo-realism, and content preservation.

We summarize our main contributions as follows:
\begin{enumerate}
    \item We introduce S2ST, an innovative diffusion-based I2IT method that enables global manipulations of complex scenes for applications requiring high content fidelity, specifically focusing on automotive-related applications. 
    \item We utilize seed translation and DDIM sampling trajectory optimization to achieve photo-realistic target domain appearance, while retaining scene structure as well as domain-agnostic low-level details.
    \item We compare our method to SoTA GAN-based and DM-based methods and present quantitative as well as qualitative evaluations and results. 
\end{enumerate}
For additional results and access to the accompanying code, please visit our project page.
%at {\color{magenta}$[URL]$}.

\section{Related Work}
\label{sec:RelatedWork}

Unpaired I2IT has garnered increasing attention and made significant progress in recent years due to its broad applicability in various computer vision and image processing tasks, such as style transfer \cite{zhu2017unpaired, liu2017unsupervised, huang2018multimodal, jiang2020tsit, dutta2022seeing}, semantic segmentation \cite{park2019semantic, guo2020gan, li2020simplified, liu2019few}, image inpainting \cite{zhao2020uctgan, song2018contextual} and image enhancement \cite{chen2018deep}. To enforce structural preservation during the translation process, most GAN-based unpaired I2IT models employ a cycle consistency mechanism \cite{zhu2017unpaired, hoffman2017cycada}, where an image is translated from the source domain to the target domain and back to the source during training.
%Often, the same translation engines are used while the source and target roles are exchanged, thus ensuring detail preservation even further. Cycle consistency accompanied by a similarity loss measured between the source (or target) and reconstructed images enables unsupervised training of the translation model.
Cycle consistency avoids mode collapse and enables unsupervised training of the translation model. While GAN-based I2IT excel in preserving details, they often fall short of capturing subtle local effects caused by differences between the domains, such as correctly turning on lights (and adding their reflections) in a day-to-night translation.

Following recent progress in diffusion-based generative models \cite{sohl2015deep,ho2020denoising,dhariwal2021diffusion,saharia2022photorealistic,ramesh2022hierarchical,rombach2022high,song2020denoising}, significant improvements have been achieved by utilizing such models for a variety of image editing and manipulation tasks. \par 

One approach in image editing involves employing masks \cite{avrahami2022blended, nichol2021glide} or cross-attention maps \cite{parmar2023zero, hertz2022prompt} to focus the editing process on specific regions of the image while leaving the remainder largely untouched. While demonstrating remarkable object editing results, these methods are less suited for global editing, where the desired changes affect the entire image.

Global edits may be achieved using SDEdit \cite{meng2021sdedit}, where an input guidance image is first corrupted with Gaussian noise and then refined using denoising. This method exhibits a trade-off between realism and faithfulness, balanced by adjusting the level of noise in the initial corruption phase. Thus, SDEdit is better suited for generating images based on rough guidance, such as a sketch or a segmentation map, and less so for fine-grained image-to-image translation.

InstructPix2Pix~\cite{brooks2023instructpix2pix} trains a model with a large collection of instruction-based image editing examples, generated using a fine-tuned LLM (GPT-3 \cite{NEURIPS2020_1457c0d6}) and a text-to-image model \cite{ramesh2022hierarchical}. While it enables image editing by following human instructions for a diverse set of edits, we demonstrate in \Cref{fig:compare} that it does not deliver satisfactory results for translations such as day-to-night.

%Pix-to-pix zero \cite{parmar2023zero} uses CLIP embedding of two sentence sets to determine the required editing direction. The authors propose to preserve structure by calculating cross-attention maps with and without the edit direction, setting the loss to be minimized as the L2 difference between the two. 
%Prompt-to-prompt \cite{hertz2022prompt} uses cross attention of the source image with values corresponding to edited text to retain structure, while enabling editing via changes in the prompt. 
Tumanyan \etal~\cite{Tumanyan_2023_CVPR} extract per-layer features and self-attention matrices from the DDIM-sampling steps of a guidance image and inject them into the corresponding steps of a generated image. This results in text-guided image translation which excels at changing the style of the guidance image, while preserving semantic layout, but is less capable of adhering to fine details.

%Imagic \cite{kawar2023imagic} allows non-rigid manipulations by first optimizing the target-related embedding $e_{tgt}$ to get $e_{opt}$, then fine-tuning the generative model to reconstruct the input image using $e_{opt}$ and finally interpolating between $e_{tgt}$ and $e_{opt}$ to generate the final editing result using the fine-tuned model.

Different methods have been suggested to control and optimize the diffusion process of a given pretrained model, some of which we utilize in our work.
ControlNet \cite{zhang2023adding} presents a method to finetune a pretrained diffusion model while adding a spatial conditioning to control the structural semantics of the synthesized image. 
%The DM-based methods above show remarkable editing for single centered objects or low-complexity images in general. But for images containing many diverse objects (as common in automotive scenes) they fail to retain fine details of input structure while meeting the subtle effects as required e.g., for day-to-night and clear-to-rain translations. 
In SeedSelect \cite{samuel2023all} the authors present a way to synthesize concepts from the ``long-tail'' of the data distribution by carefully selecting suitable generation seeds in the noise space.

Our proposed method performs a series of optimization steps along the sampling trajectory. This approach aligns with previous research in which optimization along the trajectory has also been explored. In pix2pix-zero \cite{parmar2023zero}, optimization is utilized to enforce the adherence of the inverted noise maps generated by DDIM inversion \cite{song2020denoising}
to the statistical properties of uncorrelated, Gaussian white noise. In null-text inversion \cite{mokady2023null} the optimization is employed for reconstructing the input image while facilitating the use of large guidance scale ($ > 1$). Subsequent studies, such as those presented in \cite{han2023improving, miyake2023negative} have further extended and enhanced the mechanisms proposed in \cite{mokady2023null} to achieve improved performance and computational efficiency. The goal of these methods is to enable text-driven editing of real images, while our trajectory optimization aims at detail preservation in global image-to-image translations.

%In this work, we present the first diffusion-based I2IT model designed for global editing tasks, such as day-to-night or clear-to-fog transformations, using a single pre-trained Diffusion Model. Our model is capable of operating in complex automotive scenes, guided by textual input, and only requiring a few samples from the target domain. \par
%\input{sec/3_pre}
\section{Method}
\label{sec:S2ST}

\subsection{Preliminaries}
\label{subsec:Preliminaries}
Diffusion Models (DMs) are generative models that synthesize images by iteratively denoising an initial randomly sampled noise $x_T \sim \mathcal{N}(0,I)$ in a controlled manner~\cite{sohl2015deep,ho2020denoising}. We refer to the inital noisy image $x_T$ as the diffusion seed. Conditional DMs are diffusion models trained to generate images guided by some input control signal $c$. We denote by $\epsilon_t$ the noise predicted by the denoiser $\epsilon_\theta$ at timestamp $t$, \ie, $\epsilon_t = \epsilon_\theta(x_t, t, c)$. Text-to-image diffusion models typically use Classifier-Free Guidance~\cite{ho2022classifier}, a mechanism where a guidance scale $\omega$ is used to control the amount of influence of the condition $c$ on the final output:
\begin{equation}
\label{eq:cfg}
\epsilon_t = \epsilon_\theta(x_t, t, c_\varnothing) + \omega\cdot(\epsilon_\theta(x_t, t, c) - \epsilon_\theta(x_t, t, c_\varnothing)) 
\end{equation}
where $c_\varnothing$ denotes the null-text (``'') embedding.

The term $\DDIM_\mathrm{sam}$ \cite{song2020denoising} refers to a method of synthesizing an image from a seed using a pre-trained diffusion model, with fewer denoising steps. A deterministic version of $\DDIM_\mathrm{sam}$  is shown below:
\begin{equation}
\label{eq:ddimSamp}
x_{t-1} = \sqrt{\alpha_{t-1}}\cdot \pred{x}{0} + \sqrt{1-\alpha_{t-1}}\cdot\epsilon_\theta({x_t,t,c})
\end{equation}
where $\pred{x}{0}$ is the clean image predicted by the model:
\begin{equation}
\label{eq:predx0}
\pred{x}{0} = \frac{x_t -\sqrt{1-\alpha_t}\cdot \epsilon_\theta({x_t,t,c})}{\sqrt{\alpha_t}}
\end{equation}
Here $x_t$ a noisy sample at time $t$ and $\alpha_{1:T} \in (0,1]^T$ are the diffusion schedule hyperparameters. In this paper, we utilize DDIM inversion (denoted $\DDIM_\mathrm{inv}$) to find a noise map $x_T$ that yields the input image $x_0$. A deterministic version of this process can be formulated as:
\begin{equation}
\label{eq:dir_xt}
x_{t+1} = \sqrt{\alpha_{t+1}}\cdot \pred{x}{0} + \sqrt{1-\alpha_{t+1}}\cdot\epsilon_\theta({x_t,t,c})
\end{equation}

Latent Diffusion Models (LDMs) \cite{rombach2022high} are DMs applied within a latent space of a powerful pretrained autoencoder, rather than directly on the image pixel space. In this paper we adopt Stable Diffusion \cite{rombach2022high} which encodes an input image $x_0 \in \mathcal{R}^{X\times X\times 3}$ to a latent representation $z_0\in \mathcal{R}^{S \times S \times 4}$.

\subsection{S2ST: Seed-to-Seed Translation}
\label{subsec:Our}

We translate an input image from a source domain $A$ to a target domain $B$ using two steps, as illustrated in \Cref{fig:overall}:
\begin{enumerate}
\item Seed Translation, where the seed obtained by inverting the input image from domain $A$ is translated to one that yields a generated image in domain $B$, with content and structure similar by those of the input image.
\item Trajectory Optimization, where we further enhance the structural similarity between the source image and the generated result, while preserving the appearance of the target domain.
\end{enumerate}

In this section we formulate the two aforementioned steps, and also introduce a correlation factor that represents the sensitivity of each latent channel to the differences in appearance between the source and target domains.

\subsubsection{Seed Translation}
 
Given an input image $x^A \in A$, our objective is to generate a related image $x^B$ that ideally depicts the same semantic content as $x^A$, and preserves its structure, while adopting an appearance typical of images in the target domain $B$.
Rather than attempting to perform the translation in image space, we opt to do so in the noisy latent space of a pre-trained LDM, induced by \DDIM.
Specifically, let $\za{0}$ be the latent representation of $x^A$, given by the LDM's encoder, and let $\za{T}$ be the latent seed obtained from $\za{0}$ via DDIM inversion,
$\za{T} = \DDIM_{\mathrm{inv}}(\za{0})$. We seek a translated seed, $\zb{T}$, which would yield a clean latent $\zb{0} = \DDIM_\mathrm{sam}(\LDM_\theta, \zb{T})$ that decodes into the desired target domain image $x^B$. Here $\DDIM_\mathrm{sam}$ is the deterministic version of the DDIM sampling process, and $\LDM_\theta$ is the pre-trained model.
%
%The deterministic nature of $\DDIM_\mathrm{sam}$ forces the existence of unique $\zb{T}$. Now instead of translating $\za{0}$ into $\zb{0}$ in the clean-image space, we can translate $\za{T}$ into $\zb{T}$ in the noisy seed space.
By performing the translation in the noisy seed space we leverage the powerful priors of $\LDM_\theta$ to keep $x^B$ on the natural image manifold, thereby ensuring its natural appearance.

\begin{figure*}[t]
  \centering
  %\fbox{\rule{0pt}{2in} \rule{0.9\linewidth}{0pt}}
   \includegraphics[width=0.84\linewidth]{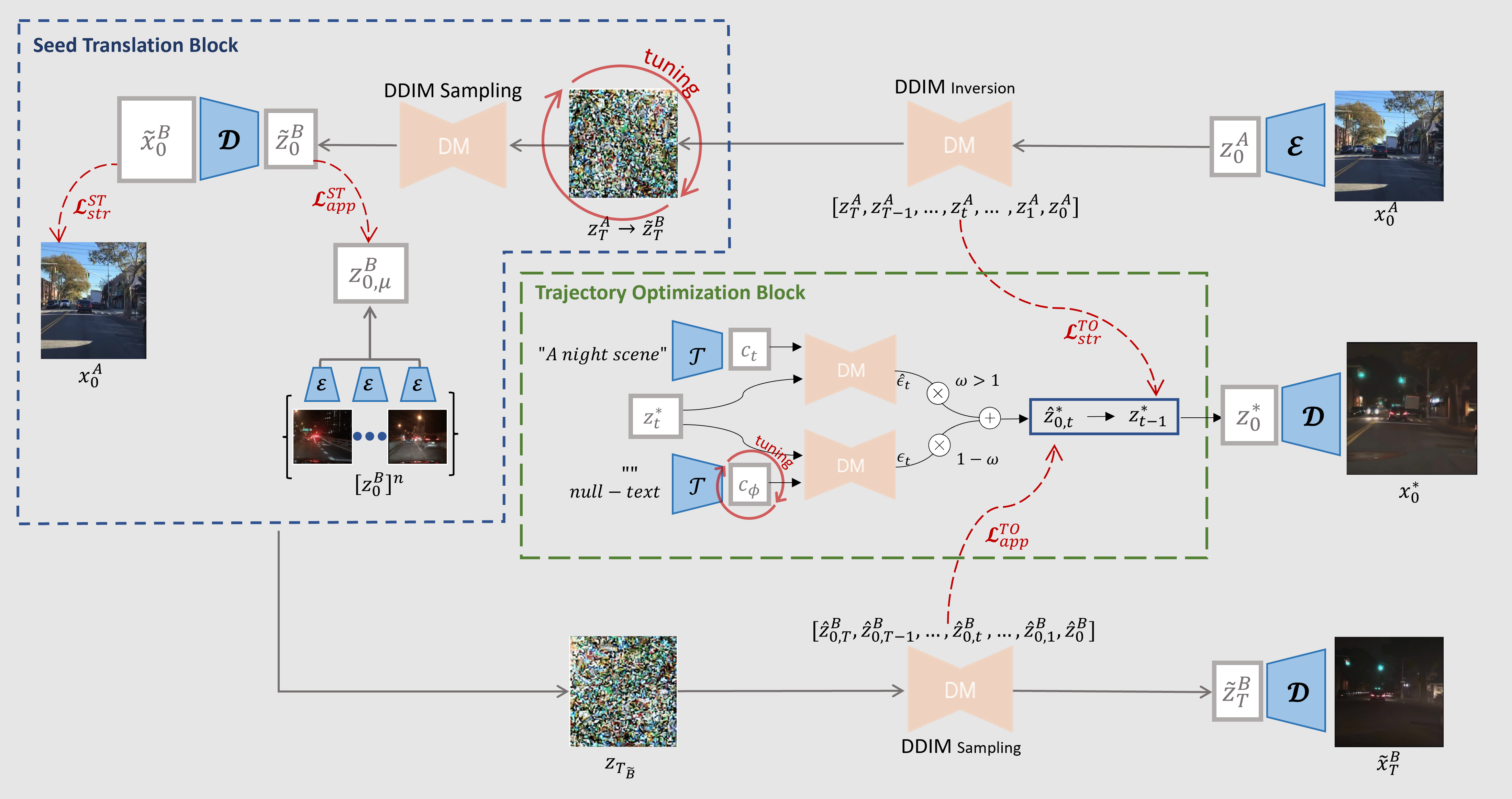}

   \caption{\textbf{S2ST framework, in detail.}  The encoding $\za{0}$ of the source image is DDIM-inverted to an initial seed $\za{T}$ and§ iteratively translated into a target domain seed $\appzb{T}$ using the Seed Translation Block. Next, starting from $\appzb{T}$, each DDIM sampling step is optimized in the Trajectory Optimization block, until the final translated latent $\zstar{0}$ is obtained and decoded.}
   \label{fig:detailed}
\end{figure*}

%Given $\za{0}$, we obtain the initial seed $\za{T}$ using $\DDIM_{\mathrm{inv}}$ (see \Cref{fig:overall}).
This translation of the initial seed $\za{T}$ into an approximation of $\zb{T}$, denoted $\appzb{T}$, is carried out using an iterative optimization process, summarized in \Cref{alg:STB} and illustrated in \Cref{fig:detailed}.
In each iteration of the process, the current seed
$\appzb{T}$ is DDIM-sampled to obtain a clean latent $\appzb{0}$. Two loss components are then used to obtain the gradients for the optimization of $\appzb{T}$: structure loss $\LossSTs$, and appearance loss $\LossSTa$, as described below. 

%For every iteration $i$ Let $z_{T_i}$ be the approximated translated seed, optimized by $i$ optimization steps. In every iteration $i$, $z_{T_i}$ is being sampled to create a ``clean" image $z_{0_i}$ using a fixed $\LDM_\theta$ guided by a target-related textual prompt. The seed is optimized to achieve similarity between $z_{0_i}$ and the input image in a structural aspects, and a batch of samples from the target domain in terms of appearance. For this purpose, we employ two loss components: structure loss $\mathcal{L}_{str,ST}$, and appearance loss $\mathcal{L}_{app,ST}$, described below. 

\textbf{Structure loss (ST phase)}: we enforce structural similarity by minimizing the distance between the Sobel gradients \cite{kanopoulos1988design} of the input image $\xa{0}$ and the decoded synthesized image $\appxb{0}$, obtained by decoding $\appzb{0}$ using the LDM's decoder. %We utilize the Sobel operator \cite{kanopoulos1988design} to extract the gradients.
The formal definition of this loss is:
 \begin{equation}
 \label{eq:str_loss}
    \LossSTs = \left\|\mathrm{Sobel}(\xa{0}) - \mathrm{Sobel}(\appxb{0})\right\|^2_2
 \end{equation}

\textbf{Appearance loss (ST phase)}: We adopt the appearance loss proposed by Samuel~\etal~\cite{samuel2023all}, where the generated latent $\appzb{0}$ is compared to the mean latent representation of $n$ samples from the target domain, denoted as $\zb{0,\mu}$. The loss function is defined as follows:
 \begin{equation}
 \label{eq:app_loss}
 \LossSTa =  \left\|\zb{0,\mu}-\appzb{0}\right\|^2_2
 \end{equation}

 The Seed Translation block is described in \Cref{alg:STB}, and illustrated in \Cref{fig:detailed}.

\begin{algorithm}
\caption{Seed Translation}\label{alg:STB}
\begin{algorithmic}
%\State \textbf{Input:} $\text{z_T_A}$
\State \textbf{Input:} $\za{T}, \za{0}, \xa{0}$
\State  $\hspace{1.3em} [\zb{0}]^n$ \Comment{\small{n ($\sim 5$) images from the target domain}}
\State  $\hspace{1.5em} c_\tau$ \Comment{\small{prompt describing the target domain}}
\State  $\hspace{1.5em} \lambda_{app}, \lambda_{str}$ \Comment{\small{loss weights}}
\State  $\hspace{1.5em} N$ \Comment{\small{$\#$ optimization steps}}

\normalsize
\State \textbf{Preprocess:} $\zb{0,\mu} = \mathrm{Mean}([\zb{0}]^n)$
\State  $\hspace{1.5em} \appzb{T} = \za{T}$ 
\State  $\hspace{1.5em} \nabla_{\!A}= \textrm{Sobel}(\xa{0})$
\For {i = 0...N}
    \State $\appzb{0} = \DDIM_\mathrm{sam}(\LDM_\theta, \appzb{T}, c_\tau)$
    \State $\appxb{0} = \VAE_{\mathrm{decode}}(\appzb{0})$
    \State $\nabla_{\!B}= \textrm{Sobel}(\appxb{0})$
    \State $\Loss = \lambda_{app}\cdot\LossSTa(\zb{0,\mu},\appzb{0}) + \lambda_{str}\cdot\LossSTs(\nabla_{\!A}, \nabla_{\!B})$
    %\State $\LossSTa$%= ||$\zb{0,\mu}$^2-z_{0_i}^2 ||$
    %\State $\LossSTs$% = ||grad_{i}^2-grad_{\mathrm{source}}^2 ||$
    %\State $\Delta z_{T_{i+1}} = \nabla_{z_{T_{i+1}}}(\lambda_{app}\cdot\mathcal{L}_{app} + \lambda_{str}\cdot\mathcal{L}_{str})$
    \State $\appzb{T} = \mathrm{UPDATE}(\appzb{T},\nabla_{\appzb{T}}\Loss)$
\EndFor
 %$z_{T_{\mathrm{N}}} \rightarrow z_{T_{\widetilde{B}}}$
\State \textbf{return} $\appzb{T}$
\end{algorithmic}
\end{algorithm}

\subsubsection{Trajectory Optimization}

The seed translation step typically results in a synthetic image that is well-aligned with the target domain, but may exhibit deviations from the source image in content and structure. In order to restore the structural similarity between the source image and the generated result, we further optimize the entire sequence of latents produced by the DDIM sampling process, starting from the translated seed $\appzb{T}$. 

Specifically, we use the \emph{latent trajectory} obtained while inverting the input image, $T_\textrm{inv} = \left[\za{T}, \za{T-1}, ... , \za{1}, \za{0}\right]$, as a reference for structure reservation. At the same time, \emph{predicted clean latents} generated by DDIM when starting from the translated seed $\appzb{T}$,  serve as a reference for target domain appearance preservation. We denote this sequence by $T_\textrm{gen} = \left[\predzb{0,T}, \predzb{0,T-1}, ... , \predzb{0,1}, \predzb{0}\right]$.

The trajectory optimization process is described in Algorithm \ref{alg:TOB}, and illustrated in \Cref{fig:detailed}. We adopt the null-text optimization mechanism of Mokady \etal~\cite{mokady2023null} to enhance the structural similarity between the synthesized image and the source, while preserving the appearance of the target domain. The null-text optimization mechanism suggests replacing the null condition embedding in the classifier-free guidance (CFG) with an optimized embedding $c_\phi$ to encode the difference between the inversion and sampling trajectories to enable image reconstruction with high CFG scale. Here we adopt the mechanism to enable reconstruction of the details from the source image, while preserving the target-domain appearance achieved during the ST phase. We control the optimization using two losses, $\LossTOs$ and $\LossTOa$, enforcing similarity in structure and appearance, respectively.

%where $\DDIM_\textrm{step}(z_{t}, t, [c_t, c_\theta])$ denotes one $\DDIM$ sampling step from $t$ to $t-1$, where the CFG $[c_t, c_\theta]$.

\begin{algorithm}
\caption{Trajectory Optimization}
\label{alg:TOB}
\begin{algorithmic}
\State \textbf{Input:} $T_\textit{inv} = [\za{T}, \za{T-1}, \dots , \za{1}, \za{0}]$
\State  $\hspace{0.5em} T_\textit{gen} = [\predzb{0,T}, \predzb{0,T-1}, ... , \predzb{0,1}, \predzb{0}]$
\State  $\hspace{0.5em} \appzb{0}$ \Comment{\small{the translated seed from the ST phase}}
\State  $\hspace{0.5em} c_\tau$ \Comment{\small{prompt describing the target domain}}
\State  $\hspace{0.5em} \eta$ \Comment{\small{target appearance correlation factor}} 
\State  $\hspace{0.5em} \lambda_{app}, \lambda_{str}$ \Comment{\small{loss weights}} \normalsize
\State \textbf{Initiate:} $c_{\phi}$
\State  $\hspace{0.5em} \zstar{T} = \appzb{T}$
\For {t = T...1}
%\State $\zstar{t} = z^*_t$
\For{j = 0...N}
    \State $\zstarpred{0,t}, \zstar{t-1} = \DDIM_\mathrm{step}(\zstar{t}, t, [c_t, c_\phi])$
    %\State $\mathcal{L}_{app} $%= ||w \cdot({\mathrm{pred}^*_{z0_{t_j}}}^2-T_{tar}[t]^2) ||$
    %\State $\mathcal{L}_{str} $%= ||{z^*_{{t-1}_j}}^2-T_{inv}[t]^2 ||$
    \State $\Loss = \lambda_{app}\cdot\LossTOa(\zstarpred{0,t},\predzb{0,t}) + \lambda_{str}\cdot\LossTOs(\zstar{t-1},\za{t-1})$
    %\State $\Delta c_{\phi} = \nabla_{c_{\phi}}(\lambda_{app}\cdot\mathcal{L}_{app} + \lambda_{str}\cdot\mathcal{L}_{str})$
    \State $c_{\phi} = \mathrm{UPDATE}(c_{\phi},\nabla_{c_{\phi}}\Loss)$
\EndFor
%\State $z^*_{t-1} = z^*_{{t-1}_j}$
\EndFor
%\State $z^*_0 \rightarrow z^*_{0_{\tilde{B}}}$
\State \textbf{return} $\zstar{0}$
\end{algorithmic}
\end{algorithm}

\textbf{Structure loss (TO phase)} We require similarity between the optimized intermediate latents $\zstar{t}$ and the corresponding latents in the inversion trajectory of the input image. Formally,
\begin{equation}
    \LossTOs = \left\|\zstar{t}-\za{t} \right\|^2_2
\end{equation}

\textbf{Appearance loss (TO phase)} 
As $T_\textrm{gen}$ starts from the translated seed $\appzb{T}$, it already consists of latents that are aligned with the target domain. To ensure that we do not deviate too far from the desired appearance, we enforce histogram similarity between the clean latent $\zstarpred{0,t}$ predicted at each timestamp $t$ and the corresponding prediction $\predzb{0,t}$ from $T_\textrm{gen}$. 
In our experiments, we observed that some latent channels are more strongly correlated with the final appearance than others. We therefore weight the histogram similarity with a domain related channel-wise correlation factor $\eta$, which reflects the sensitivity of each latent channel to the difference in appearance between domains $A$ and $B$ (a detailed explanation is provided in the supplementary material). Formally, the appearance loss is defined as:

\begin{equation}
    \LossTOa = \sum_{i=1}^n
    \eta_i \cdot \left\|h(\zstarpred{0,t})_i-h(\predzb{0,t})_i \right\|^2_2
\end{equation}
where $h(\cdot)$ denotes the histogram operator, and $n$ is the number of channels in the latent representation.
\section{Experiments}
\label{sec:exps}

%\subsection{Implementation Details}
\label{subsec:Impl}

We implemented our S2ST method using a pre-trained Stable Diffusion v2.1 model. We first fine-tuned the model to generate automotive scenes with a random subset of 50k images from the BDD100k~\cite{yu2020bdd100k} training set (containing mixed domains: day, night, rain, fog).
Specifically, we train a ControlNet \cite{zhang2023adding} with simple Canny maps as the spatial conditioning controls. In the S2ST process we use 10 ST iterations followed by 10 TO iterations at each DDIM step. For both optimizations, we use the ADAM optimizer~\cite{kingma2014adam}. Both the DDIM-inversion and sampling processes perform 20 DDIM steps. Additional implementation details are provided in the supplementary material.

We compare S2ST to leading GAN-based I2IT baselines. While GAN-based methods dedicated for global unpaired I2IT of complex (automotive) scenes are prevalent, dedicated diffusion-based solutions are currently absent in the literature. Consequently, our quantitative comparisons focus on GAN-based methods. However, in our qualitatitve comparison in \Cref{fig:compare}, we also include a recent diffusion-based I2IT approach, namely Instruct-Pix2Pix~\cite{brooks2023instructpix2pix}. Additional comparisons with diffusion-based methods are included in the supplementary material.

\subsection{Quantitative Comparison}
\label{subsec:quan}

We quantitatively compare our method to CycleGAN~\cite{zhu2017unpaired}, MUNIT~\cite{huang2018multimodal}, TSIT~\cite{jiang2020tsit} and AU-GAN~\cite{kwak2021adverse} using the BDD100k~\cite{yu2020bdd100k} dataset. As these methods necessitate individual training for each specific source-target domain pair, our quantitative evaluation was centered on the prevalent use-case of day-to-night translation. For each model, we used the default provided hyper-parameters.

We employ a standard evaluation protocol commonly used in prior GAN-based I2IT works \cite{brock2018large, liang2021high, liu2019learning} for quantitative evaluation of the day-to-night translation task. We adopt the Kernel Inception Distance (KID)~\cite{binkowski2018demystifying} to measure the realism of the generated images in the target domain. We chose KID over Fréchet Inception Distance (FID)~\cite{heusel2017gans}, since we used relatively small sets of synthetic images (4k). 
We measured  the structural similarity using the Structural Similarity Index Metric (SSIM)~\cite{wang2004image}. As SSIM also contains components affected by photometric aspects (\eg, luminance and contrast), we further measured the structural similarity using Dino \cite{caron2021emerging} features, as proposed by Tumanyan \etal~\cite{tumanyan2022splicing} (Struct-dist). 

\begin{table}
    \centering
    \begin{tabular}{c|c|c|c|c}
         Method &  KID $\downarrow$
&  SSIM $\uparrow$& Struct-dist $\downarrow$& \\
         \hline
         \textbf{CycleGAN}&  22.554&  0.506& \textcolor{red}{0.067}&\\
         \textbf{MUNIT}&  \textcolor{red}{22.508}&  0.470& 0.074&\\
         \textbf{TSIT}&  28.208&  0.525& \textcolor{blue}{0.057}&\\
         \textbf{AU-GAN}&  23.011&  \textcolor{red}{0.558}& 0.072&\\
         \textbf{S2ST(ours)}&  \textcolor{blue}{21.234}&  \textcolor{blue}{0.590}& \textcolor{red}{0.067}&\\
        \hline
    \end{tabular}
    \caption{\textbf{Quantitative results.} Day-to-Night translation over BDD100k~\cite{yu2020bdd100k}. For each metric, top and second scores are colored \textcolor{blue}{blue} and \textcolor{red}{red}, respectively}
    \label{tab:quant1}
\end{table}

The results, reported in \Cref{tab:quant1}, show that our model outperforms the GAN-based competitors in terms of KID and SSIM, while maintaining a competitive level in Struct-dist. These results are aligned with our goal to achieve better appearance in the target domain while maintaining the level of content preservation compared to GANs.

\subsection{Human Evaluation}
\label{subsec:qual}

We further evaluated S2ST's performance via a human perceptual evaluation study, where the participants were shown pairs of synthetic night images, translated from the same input (day) image by different methods (ours, and a baseline from one of: CycleGAN \cite{zhu2017unpaired}, MUNIT \cite{huang2018multimodal}, TSIT \cite{jiang2020tsit}, or AU-GAN \cite{kwak2021adverse}). It is worth re-emphasizing that we employed default hyperparameters without any specific tuning. In MUNIT, as we did not find an available configuration for day-to-night, we used hyperparameters originally proposed for Cityscapes-to-SYNTHIA translation.
The participants were asked to answer two questions for each pair of images: 

\begin{enumerate}
    \item Which image looks more like a natural night image?
    \item (given the source image) Which image looks more like a natural night image preserving the content of the source image?
\end{enumerate}

\begin{figure}[h!]
  \centering
   \includegraphics[width=0.8\linewidth]{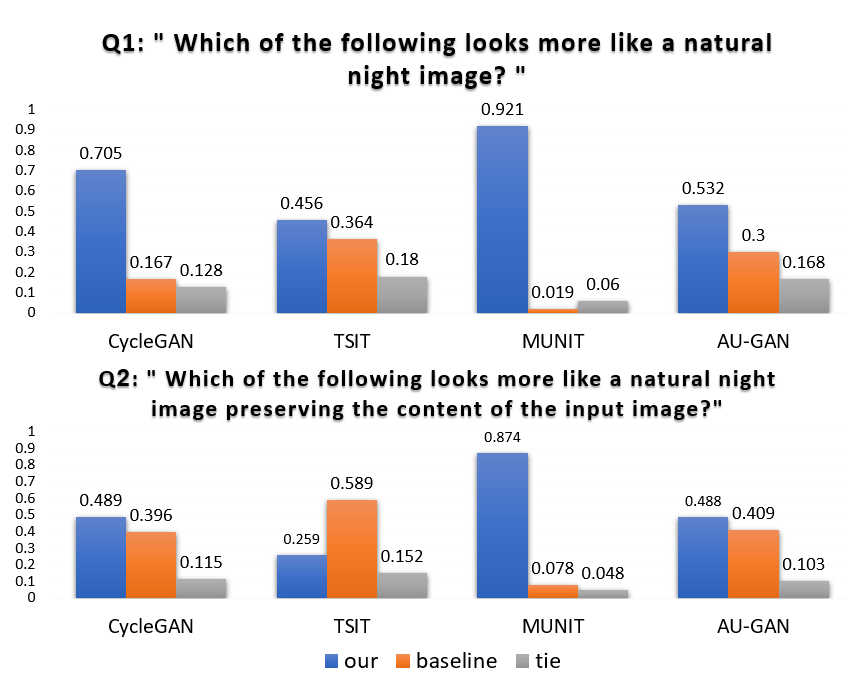}

   \caption{\textbf{User study results:} Summary of responses to the two questions asked in the user study. The results show that our method outperforms all baselines in terms of target domain appearance (Q1), and all but one in terms of content preservation (Q2).}
   \label{fig:MOS}
\end{figure}

The samples where randomly drawn from a pool of 4K source (day) images randomly sampled from BDD100k.
In total, we collected $1300$ answers, with the results summarized in \Cref{fig:MOS}. As can be seen, evaluators exhibit a strong preference towards our method in terms of achieving the target domain appearance (Q1), while preserving high fidelity to the content of the source image (Q2). It should be noted that one baseline, TSIT \cite{jiang2020tsit}, outperformed our model in terms of content preservation, as evident from the structural distinctiveness (Struct-dist) scores presented in \Cref{tab:quant1}. We provide a detailed discussion of this comparison in \Cref{sec:disc} below. \par

%\begin{figure}[h!]
%    \centering
%    \begin{subfigure}{0.4\textwidth}
%        \centering
%        \includegraphics[width=\textwidth]{figures/MOS_Q1.png}
%    \end{subfigure}

%    \begin{subfigure}{0.4\textwidth}
%        \centering
       % \includegraphics[width=\textwidth]{figures/MOS_Q2.png}
    %\end{subfigure}

    %\caption{\textbf{User study results:} Summary of responses to the two questions asked in the user study. The results show that our method outperforms all baselines in terms of target domain appearance (Q1), and all but one in terms of content preservation (Q2).}
    %\label{fig:MOS}
%\end{figure}

\begin{figure}[t]
    \centering
     \includegraphics[width=0.8\linewidth]{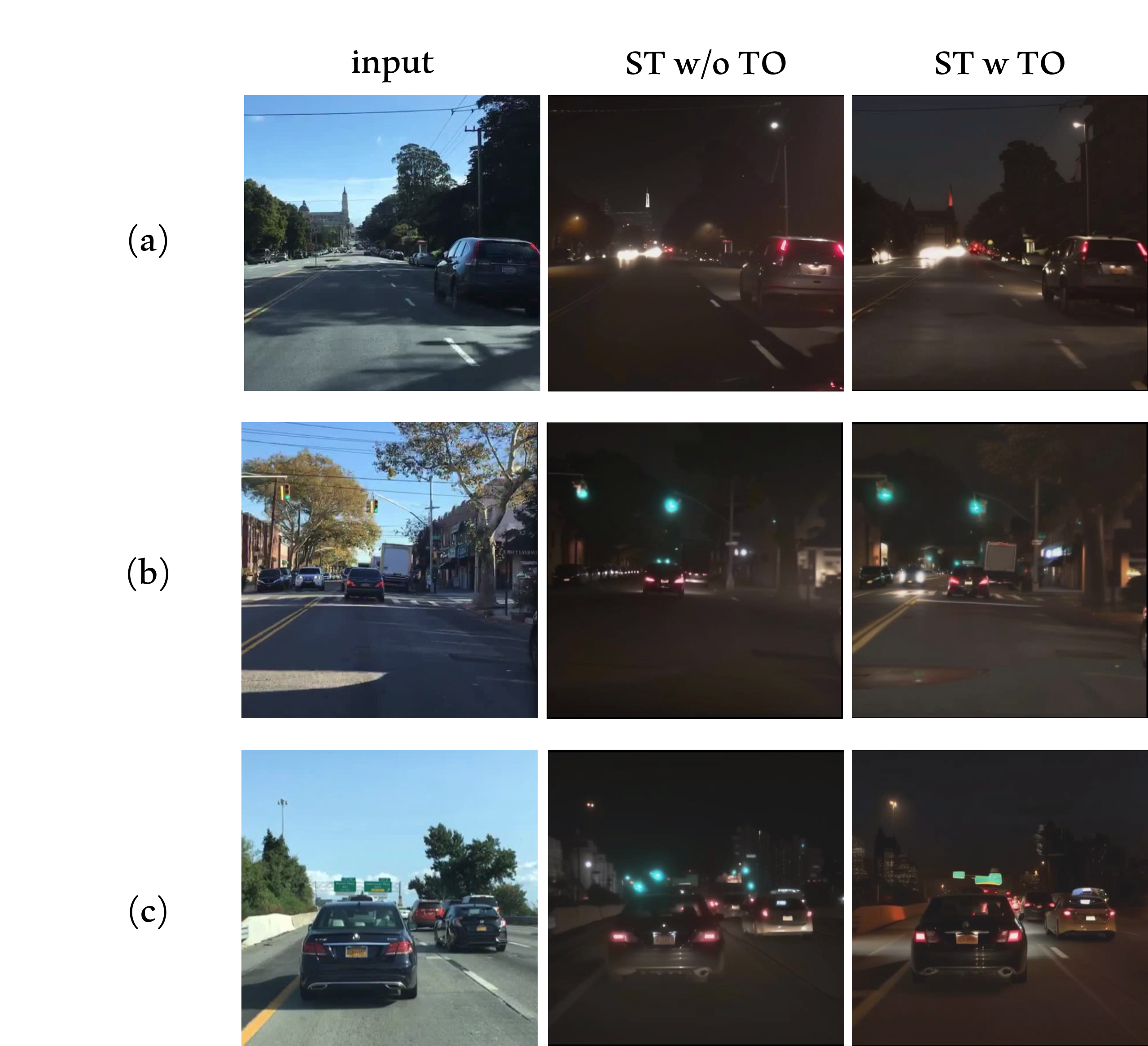}
  
     \caption{\textbf{Effect of the two components of our model:} Seed Translation (ST) and Trajectory Optimization (TO). While the main domain shift is performed during the ST phase, the TO phase enhances the image and the preservation of details with respect to the input image.}
     \label{fig:ablation}
     \vspace{-1ex}
\end{figure}

Finally, we ablate the second phases of our method (TO) 
to examine its effectiveness. Figure \ref{fig:ablation} shows two translated versions for several input images (left), with ST only (middle), and with both ST and TO (right). It is noticeable that the ST phase successfully transforms the image to exhibit night-time appearance in a global manner, the subsequent trajectory optimization (TO) further enhances the details in the image and the effects related to those details. Images synthesized using the TO phase exhibit increased sharpness and are more photorealistic. In many instances, content details omitted during the ST phase are effectively restored during the TO phase, as seen in the ``horizon line" in \Cref{fig:ablation}(a) or the restoration of the left front-facing car in \ref{fig:ablation}(b). Additionally, the TO phase amplifies semantics-related effects of the target domain, exemplified by the enhanced projection of hidden headlights in \Cref{fig:ablation}(c).

More examples and additional user study details are provided in the supplementary material.

\begin{figure*}[t]
  \centering
   \includegraphics[width=0.98\linewidth]{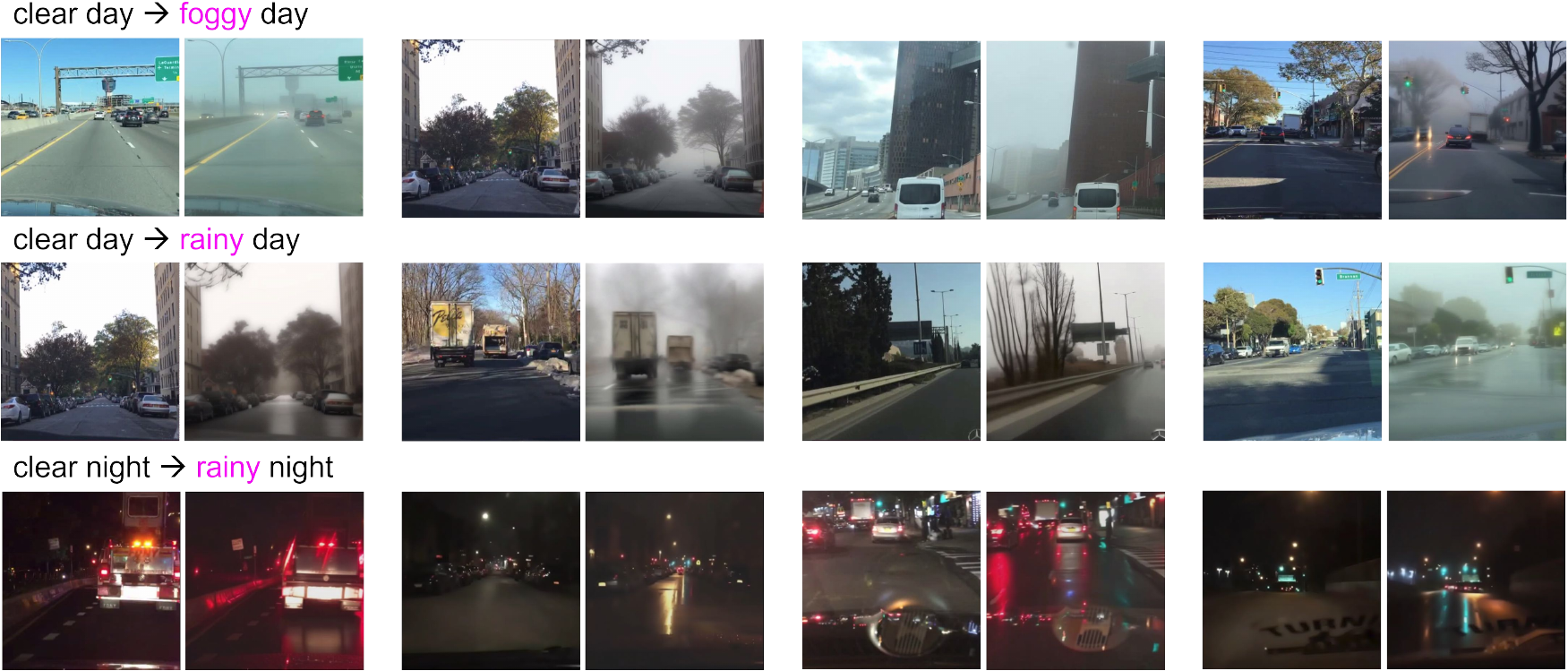}

   \caption{\textbf{Examples of S2ST results} for different source-target domain pairs. For each image pair, we show the image before and after the translation. Our translations capture the target domain appearance while preserving the structure and content of the source image.}
   \label{fig:domains}
\end{figure*}

\section{Discussion}
\label{sec:disc}

As is evident from the results presented in \Cref{sec:exps}, our model outperforms existing GAN-based methods in classic I2IT tasks with respect to target domain appearance. Moreover, our approach allows multi-domain translation, where the same diffusion model is used to translate different source domains to different target domains, unlike GAN-based methods that require training a dedicated model per source-target domain pair. Several examples of translations between different domain pairs are presented in \Cref{fig:domains}. 

We observe that the most substantial advantage of our model lies in addressing multiple target domain effects, often subtle, while remaining on the manifold of natural images. 
These capabilities indicate scene comprehension and encompass effects such as activating visible and hidden light sources in day-to-night translation, faithfully rendering reflections and scattering effects in clean-to-rain translation, and effectively handling fog effects related to depth in clean-to-fog translation (see examples in \Cref{fig:domains}).

\begin{figure*}[t]
  \centering
   \includegraphics[width=1.0\linewidth]{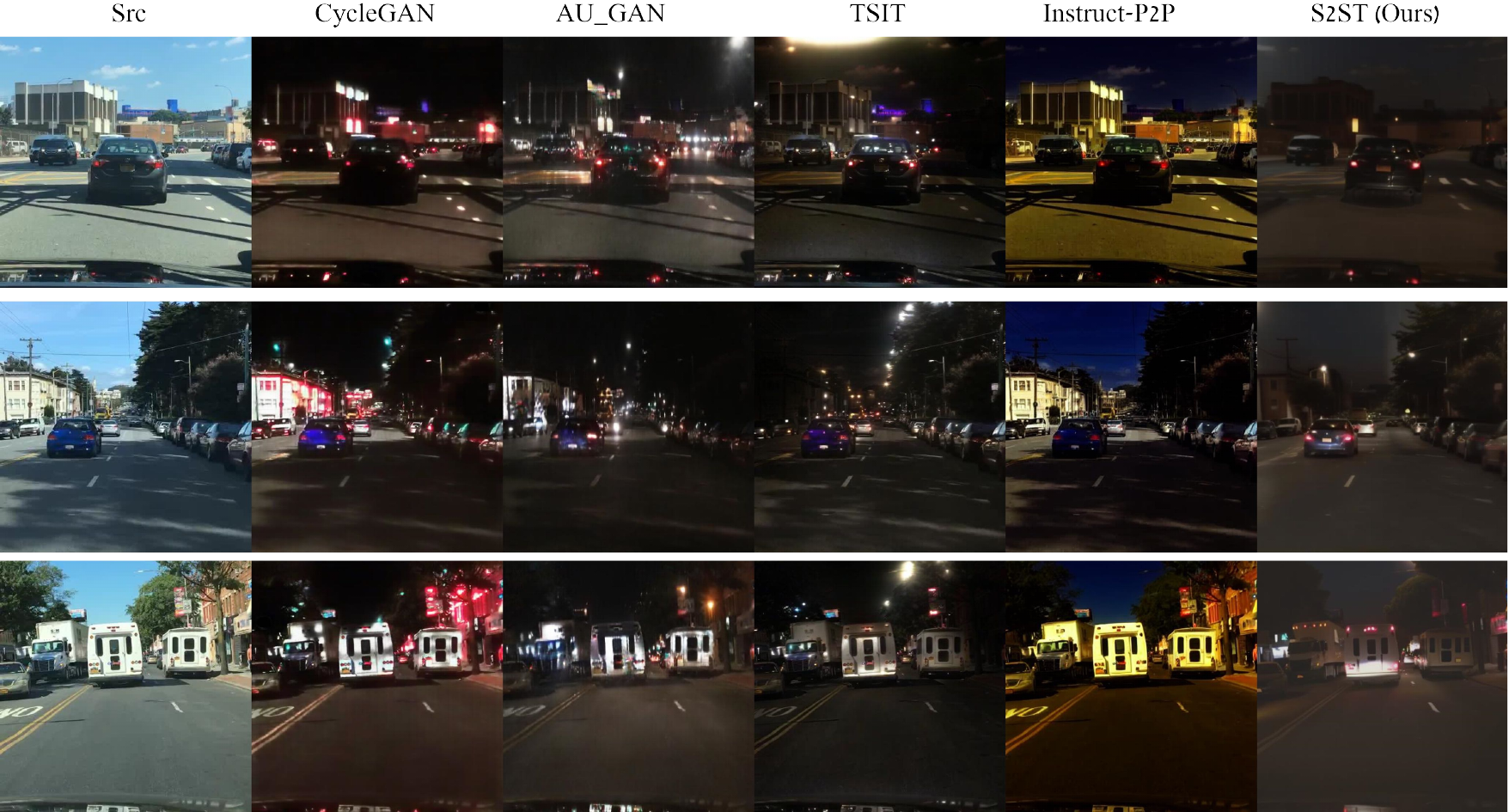}
   \caption{\label{fig:compare}
   \textbf{A visual comparison.} We compare several day-to-night results of our method with those of several GAN-based baselines: CycleGAN~\cite{zhu2017unpaired}, AU-GAN~\cite{kwak2021adverse}, TSIT~\cite{jiang2020tsit}, as well as a diffusion-based method Instruct-Pix2Pix~\cite{brooks2023instructpix2pix}. Note that the results produced by all baselines tend to retain some illumination effects from the source domain (\eg, shadows on the road, some bright surfaces), while our method modifies the illumination to better correspond to night time. GAN-based methods also tend to introduce spurious light sources, which are not necessarily located at the same locations as the light sources in the source image, while Instruct-Pix2Pix does not ``turn on'' enough light sources. Our method on the other hand, handles light sources in a more realistic manner.
   }
\end{figure*}

In GAN-based methods, the capability to replicate the characteristics of the desired domain relies on the effectiveness of a discriminative procedure. Consequently, GANs often tend to concentrate on an uncontrolled subset of salient target domain features, thereby hindering their ability to fully learn the domain subtleties. 
For instance, in Day-to-Night translations, GAN models may prioritize reducing scene luminance but may not meticulously address the lighting of potential light sources. This oversight can lead to the emergence of ``sparkles" that do not necessarily correspond to the locations of potential light sources in the daytime scene, as demonstrated in \Cref{fig:compare}. DMs, On the other hand, learn to guide a given signal towards the target manifold across various levels of noise levels, thereby compelling the model to assimilate a broad spectrum of manifold features.

With respect to the capacity to maintain structure and fine content details of the source image, we discern an inherent trade-off between the requirement to respect target domain attributes and the need to remain faithful to source content. Analyzing the user-study results presented in \Cref{fig:MOS}, we observe that the presence of effects arising from the gap between the source and target domains (\eg, increased blurriness, light sources, auto white balance and chromatic adaptation effects, and reduced visibility of dark or distant objects, all associated with day-to-night translations) inherently reduces the apparent fidelity. GAN-based models often utilize cycle consistency (\eg, day-to-night-to-day) to ensure faithfulness to the source, while DMs do not offer a mechanism controlling the drift from the source. Our approach overcomes this inherent downside by trajectory optimization that simultaneously minimizes the drift from the source's content and the target's appearance at every diffusion step.

The inquiry into which synthetic image is better involves a trade-off between unnatural artifacts, unfaithfulness to the source, and effects which do not respect the target domain. Weighting the trade-off elements and ranking the methods accordingly depends on the application for which the method is used. The ambiguity was acknowledged by some participants in the user study, nonetheless, we believe that the score differences are sufficiently significant to indicate the value of our approach.  

\subsection{Limitations and Future Work}
\label{subsec:future}

One primary shortcoming of our method is the computational cost associated with performing backpropagation through the entire sampling process (20 DDIM steps). The generation of a single sample with 10 Seed-Translation steps and 10 Trajectory-Optimization steps may consume $\sim\!170$ seconds on a single Tesla A100 GPU and $\sim\!45$ GB of GPU memory. One way to reduce the computational cost is by using a thinner DM backbone (\eg, omitting the fine-tuning over a trainable copy suggested in ControlNet \cite{zhang2023adding}). 

Yet a significant reduction of computational resources might require replacing the backpropagation scheme by an equivalent optimization algorithm. 
%\dani{A bit too vague}
In conjunction with the proposed efforts to optimize the computational processes, future work might enhance the results attained by the model. While GAN based solutions harness the cycle-consistency mechanism to enforce content preservation, our model uses the combination of inverted seed, structural loss and trajectory optimization to do the very same. All the aforementioned endeavors to preserve content within the diffusion model inherently limit its innate capacity to generate natural images on demand, as they compel it to strike a balance between generation and preservation priorities. Recent works \cite{xu2023cyclenet, wang2022diffusion} suggest using the cycle-consistency mechanism in diffusion-based methods. Either way, in order to use the full power of diffusion models as generative engines, it should contain a strong content preservation mechanism.
\section{Conclusions}
\label{sec:conc}

%We propose S2ST, a novel I2IT method that performs diffusion-based image translation of complex (\eg, automotive) scenes, guided by a simple textual prompt and a few samples typical of the target domain. \par
%Comparisons to GAN-based and diffusion-based baselines demonstrate superior performance in term of appearance in the target domain, while preserving the content of the source image. Moreover, contrary to typical GAN-based I2IT methods, our approach handles diverse domains (source and target) without requiring the training of a task-specific network for each domain pair.

We introduced S2ST, a novel method for unpaired I2IT, with a particular focus on complex automotive scenes. Our method leverages the powerful priors of diffusion models, for translating images between varied domains while adhering to their structure and content. We have shown that S2ST outperforms existing methods in terms of maintaining photo-realism and content fidelity in scenarios where the scene is complex and detail preservation is crucial. %This suggests that diffusion model-based approaches like ours could offer alternative pathways in high-fidelity image translation tasks.

In a more general context, we believe that our seed and trajectory optimization approach offers a promising tool for controlled image manipulation using diffusion. Thus, our work contributes to the evolving landscape of image manipulation techniques, underscoring the potential of diffusion models in complex image translation and editing tasks.

{
    \small
    \bibliographystyle{ieeenat_fullname}
    \bibliography{main}

\begin{thebibliography}{51}
\providecommand{\natexlab}[1]{#1}
\providecommand{\url}[1]{\texttt{#1}}
\expandafter\ifx\csname urlstyle\endcsname\relax
  \providecommand{\doi}[1]{doi: #1}\else
  \providecommand{\doi}{doi: \begingroup \urlstyle{rm}\Url}\fi

\bibitem[Avrahami et~al.(2022)Avrahami, Lischinski, and
  Fried]{avrahami2022blended}
Omri Avrahami, Dani Lischinski, and Ohad Fried.
\newblock Blended diffusion for text-driven editing of natural images.
\newblock In \emph{Proceedings of the IEEE/CVF Conference on Computer Vision
  and Pattern Recognition}, pages 18208--18218, 2022.

\bibitem[Bi{\'n}kowski et~al.(2018)Bi{\'n}kowski, Sutherland, Arbel, and
  Gretton]{binkowski2018demystifying}
Miko{\l}aj Bi{\'n}kowski, Danica~J Sutherland, Michael Arbel, and Arthur
  Gretton.
\newblock Demystifying {MMD} {GANs}.
\newblock \emph{arXiv preprint arXiv:1801.01401}, 2018.

\bibitem[Brock et~al.(2018)Brock, Donahue, and Simonyan]{brock2018large}
Andrew Brock, Jeff Donahue, and Karen Simonyan.
\newblock Large scale {GAN} training for high fidelity natural image synthesis.
\newblock \emph{arXiv preprint arXiv:1809.11096}, 2018.

\bibitem[Brooks et~al.(2023)Brooks, Holynski, and
  Efros]{brooks2023instructpix2pix}
Tim Brooks, Aleksander Holynski, and Alexei Efros.
\newblock {InstructPix2Pix}: Learning to follow image editing instructions.
\newblock In \emph{Proc.~CVPR}, pages 18392--18402, 2023.

\bibitem[Brown et~al.(2020)Brown, Mann, Ryder, Subbiah, Kaplan, Dhariwal,
  Neelakantan, Shyam, Sastry, Askell, Agarwal, Herbert-Voss, Krueger, Henighan,
  Child, Ramesh, Ziegler, Wu, Winter, Hesse, Chen, Sigler, Litwin, Gray, Chess,
  Clark, Berner, McCandlish, Radford, Sutskever, and
  Amodei]{NEURIPS2020_1457c0d6}
Tom Brown, Benjamin Mann, Nick Ryder, Melanie Subbiah, Jared~D Kaplan, Prafulla
  Dhariwal, Arvind Neelakantan, Pranav Shyam, Girish Sastry, Amanda Askell,
  Sandhini Agarwal, Ariel Herbert-Voss, Gretchen Krueger, Tom Henighan, Rewon
  Child, Aditya Ramesh, Daniel Ziegler, Jeffrey Wu, Clemens Winter, Chris
  Hesse, Mark Chen, Eric Sigler, Mateusz Litwin, Scott Gray, Benjamin Chess,
  Jack Clark, Christopher Berner, Sam McCandlish, Alec Radford, Ilya Sutskever,
  and Dario Amodei.
\newblock Language models are few-shot learners.
\newblock In \emph{Advances in Neural Information Processing Systems}, pages
  1877--1901. Curran Associates, Inc., 2020.

\bibitem[Caron et~al.(2021)Caron, Touvron, Misra, J{\'e}gou, Mairal,
  Bojanowski, and Joulin]{caron2021emerging}
Mathilde Caron, Hugo Touvron, Ishan Misra, Herv{\'e} J{\'e}gou, Julien Mairal,
  Piotr Bojanowski, and Armand Joulin.
\newblock Emerging properties in self-supervised vision transformers.
\newblock In \emph{Proceedings of the IEEE/CVF international conference on
  computer vision}, pages 9650--9660, 2021.

\bibitem[Chen et~al.(2018)Chen, Wang, Kao, and Chuang]{chen2018deep}
Yu-Sheng Chen, Yu-Ching Wang, Man-Hsin Kao, and Yung-Yu Chuang.
\newblock Deep photo enhancer: Unpaired learning for image enhancement from
  photographs with {GANs}.
\newblock In \emph{Proceedings of the IEEE conference on computer vision and
  pattern recognition}, pages 6306--6314, 2018.

\bibitem[Dhariwal and Nichol(2021)]{dhariwal2021diffusion}
Prafulla Dhariwal and Alexander Nichol.
\newblock Diffusion models beat {GANs} on image synthesis.
\newblock \emph{Advances in neural information processing systems},
  34:\penalty0 8780--8794, 2021.

\bibitem[Dutta(2022)]{dutta2022seeing}
Ujjal~Kr Dutta.
\newblock Seeing objects in dark with continual contrastive learning.
\newblock In \emph{European Conference on Computer Vision}, pages 286--302.
  Springer, 2022.

\bibitem[Guo et~al.(2020)Guo, Wang, Yang, Lv, Liu, Wu, and Huang]{guo2020gan}
Xi Guo, Zhicheng Wang, Qin Yang, Weifeng Lv, Xianglong Liu, Qiong Wu, and Jian
  Huang.
\newblock {GAN}-based virtual-to-real image translation for urban scene
  semantic segmentation.
\newblock \emph{Neurocomputing}, 394:\penalty0 127--135, 2020.

\bibitem[Han et~al.(2023)Han, Wen, Chen, Zhang, Song, Ren, Gao, Chen, Liu,
  Zhangli, et~al.]{han2023improving}
Ligong Han, Song Wen, Qi Chen, Zhixing Zhang, Kunpeng Song, Mengwei Ren,
  Ruijiang Gao, Yuxiao Chen, Di Liu, Qilong Zhangli, et~al.
\newblock Improving negative-prompt inversion via proximal guidance.
\newblock \emph{arXiv preprint arXiv:2306.05414}, 2023.

\bibitem[Hertz et~al.(2022)Hertz, Mokady, Tenenbaum, Aberman, Pritch, and
  Cohen-Or]{hertz2022prompt}
Amir Hertz, Ron Mokady, Jay Tenenbaum, Kfir Aberman, Yael Pritch, and Daniel
  Cohen-Or.
\newblock Prompt-to-prompt image editing with cross attention control.
\newblock \emph{arXiv preprint arXiv:2208.01626}, 2022.

\bibitem[Heusel et~al.(2017)Heusel, Ramsauer, Unterthiner, Nessler, and
  Hochreiter]{heusel2017gans}
Martin Heusel, Hubert Ramsauer, Thomas Unterthiner, Bernhard Nessler, and Sepp
  Hochreiter.
\newblock {GANs} trained by a two time-scale update rule converge to a local
  {Nash} equilibrium.
\newblock \emph{Advances in neural information processing systems}, 30, 2017.

\bibitem[Ho and Salimans(2022)]{ho2022classifier}
Jonathan Ho and Tim Salimans.
\newblock Classifier-free diffusion guidance.
\newblock \emph{arXiv preprint arXiv:2207.12598}, 2022.

\bibitem[Ho et~al.(2020)Ho, Jain, and Abbeel]{ho2020denoising}
Jonathan Ho, Ajay Jain, and Pieter Abbeel.
\newblock Denoising diffusion probabilistic models.
\newblock \emph{Advances in neural information processing systems},
  33:\penalty0 6840--6851, 2020.

\bibitem[Hoffman et~al.(2018)Hoffman, Tzeng, Park, Zhu, Isola, Saenko, Efros,
  and Darrell]{hoffman2017cycada}
Judy Hoffman, Eric Tzeng, Taesung Park, Jun-Yan Zhu, Phillip Isola, Kate
  Saenko, Alexei Efros, and Trevor Darrell.
\newblock {CyCADA}: Cycle-consistent adversarial domain adaptation.
\newblock In \emph{Proc.~ICLR}, 2018.

\bibitem[Huang et~al.(2018)Huang, Liu, Belongie, and
  Kautz]{huang2018multimodal}
Xun Huang, Ming-Yu Liu, Serge Belongie, and Jan Kautz.
\newblock Multimodal unsupervised image-to-image translation.
\newblock In \emph{Proceedings of the European conference on computer vision
  (ECCV)}, pages 172--189, 2018.

\bibitem[Jiang et~al.(2020)Jiang, Zhang, Huang, Liu, Shi, and
  Loy]{jiang2020tsit}
Liming Jiang, Changxu Zhang, Mingyang Huang, Chunxiao Liu, Jianping Shi, and
  Chen~Change Loy.
\newblock {TSIT}: A simple and versatile framework for image-to-image
  translation.
\newblock In \emph{Computer Vision--ECCV 2020: 16th European Conference,
  Glasgow, UK, August 23--28, 2020, Proceedings, Part III 16}, pages 206--222.
  Springer, 2020.

\bibitem[Kanopoulos et~al.(1988)Kanopoulos, Vasanthavada, and
  Baker]{kanopoulos1988design}
Nick Kanopoulos, Nagesh Vasanthavada, and Robert~L Baker.
\newblock Design of an image edge detection filter using the {Sobel} operator.
\newblock \emph{IEEE Journal of solid-state circuits}, 23\penalty0
  (2):\penalty0 358--367, 1988.

\bibitem[Kingma and Ba(2014)]{kingma2014adam}
Diederik~P Kingma and Jimmy Ba.
\newblock Adam: A method for stochastic optimization.
\newblock \emph{arXiv preprint arXiv:1412.6980}, 2014.

\bibitem[Kwak et~al.(2021)Kwak, Jin, Li, Yoon, Kim, and Ko]{kwak2021adverse}
Jeong-gi Kwak, Youngsaeng Jin, Yuanming Li, Dongsik Yoon, Donghyeon Kim, and
  Hanseok Ko.
\newblock Adverse weather image translation with asymmetric and
  uncertainty-aware {GAN}.
\newblock \emph{arXiv preprint arXiv:2112.04283}, 2021.

\bibitem[Li et~al.(2020)Li, Cao, Jiao, Wu, and Wong]{li2020simplified}
Rui Li, Wenming Cao, Qianfen Jiao, Si Wu, and Hau-San Wong.
\newblock Simplified unsupervised image translation for semantic segmentation
  adaptation.
\newblock \emph{Pattern Recognition}, 105:\penalty0 107343, 2020.

\bibitem[Liang et~al.(2021)Liang, Zeng, and Zhang]{liang2021high}
Jie Liang, Hui Zeng, and Lei Zhang.
\newblock High-resolution photorealistic image translation in real-time: A
  {Laplacian} pyramid translation network.
\newblock In \emph{Proceedings of the IEEE/CVF Conference on Computer Vision
  and Pattern Recognition}, pages 9392--9400, 2021.

\bibitem[Liu et~al.(2017)Liu, Breuel, and Kautz]{liu2017unsupervised}
Ming-Yu Liu, Thomas Breuel, and Jan Kautz.
\newblock Unsupervised image-to-image translation networks.
\newblock \emph{Advances in neural information processing systems}, 30, 2017.

\bibitem[Liu et~al.(2019{\natexlab{a}})Liu, Huang, Mallya, Karras, Aila,
  Lehtinen, and Kautz]{liu2019few}
Ming-Yu Liu, Xun Huang, Arun Mallya, Tero Karras, Timo Aila, Jaakko Lehtinen,
  and Jan Kautz.
\newblock Few-shot unsupervised image-to-image translation.
\newblock In \emph{Proceedings of the IEEE/CVF international conference on
  computer vision}, pages 10551--10560, 2019{\natexlab{a}}.

\bibitem[Liu et~al.(2019{\natexlab{b}})Liu, Yin, Shao, Wang,
  et~al.]{liu2019learning}
Xihui Liu, Guojun Yin, Jing Shao, Xiaogang Wang, et~al.
\newblock Learning to predict layout-to-image conditional convolutions for
  semantic image synthesis.
\newblock \emph{Advances in Neural Information Processing Systems}, 32,
  2019{\natexlab{b}}.

\bibitem[Meng et~al.(2021)Meng, He, Song, Song, Wu, Zhu, and
  Ermon]{meng2021sdedit}
Chenlin Meng, Yutong He, Yang Song, Jiaming Song, Jiajun Wu, Jun-Yan Zhu, and
  Stefano Ermon.
\newblock {SDEdit}: Guided image synthesis and editing with stochastic
  differential equations.
\newblock \emph{arXiv preprint arXiv:2108.01073}, 2021.

\bibitem[Miyake et~al.(2023)Miyake, Iohara, Saito, and
  Tanaka]{miyake2023negative}
Daiki Miyake, Akihiro Iohara, Yu Saito, and Toshiyuki Tanaka.
\newblock Negative-prompt inversion: Fast image inversion for editing with
  text-guided diffusion models.
\newblock \emph{arXiv preprint arXiv:2305.16807}, 2023.

\bibitem[Mokady et~al.(2023)Mokady, Hertz, Aberman, Pritch, and
  Cohen-Or]{mokady2023null}
Ron Mokady, Amir Hertz, Kfir Aberman, Yael Pritch, and Daniel Cohen-Or.
\newblock Null-text inversion for editing real images using guided diffusion
  models.
\newblock In \emph{Proceedings of the IEEE/CVF Conference on Computer Vision
  and Pattern Recognition}, pages 6038--6047, 2023.

\bibitem[Mou et~al.(2023)Mou, Wang, Xie, Zhang, Qi, Shan, and
  Qie]{mou2023t2iadapter}
Chong Mou, Xintao Wang, Liangbin Xie, Jian Zhang, Zhongang Qi, Ying Shan, and
  Xiaohu Qie.
\newblock {T2I-Adapter:} learning adapters to dig out more controllable ability
  for text-to-image diffusion models.
\newblock \emph{arXiv preprint arXiv:2302.08453}, 2023.

\bibitem[Nichol et~al.(2021)Nichol, Dhariwal, Ramesh, Shyam, Mishkin, McGrew,
  Sutskever, and Chen]{nichol2021glide}
Alex Nichol, Prafulla Dhariwal, Aditya Ramesh, Pranav Shyam, Pamela Mishkin,
  Bob McGrew, Ilya Sutskever, and Mark Chen.
\newblock {GLIDE}: Towards photorealistic image generation and editing with
  text-guided diffusion models.
\newblock \emph{arXiv preprint arXiv:2112.10741}, 2021.

\bibitem[Park et~al.(2019)Park, Liu, Wang, and Zhu]{park2019semantic}
Taesung Park, Ming-Yu Liu, Ting-Chun Wang, and Jun-Yan Zhu.
\newblock Semantic image synthesis with spatially-adaptive normalization.
\newblock In \emph{Proceedings of the IEEE/CVF conference on computer vision
  and pattern recognition}, pages 2337--2346, 2019.

\bibitem[Parmar et~al.(2023)Parmar, Kumar~Singh, Zhang, Li, Lu, and
  Zhu]{parmar2023zero}
Gaurav Parmar, Krishna Kumar~Singh, Richard Zhang, Yijun Li, Jingwan Lu, and
  Jun-Yan Zhu.
\newblock Zero-shot image-to-image translation.
\newblock In \emph{ACM SIGGRAPH 2023 Conference Proceedings}, pages 1--11,
  2023.

\bibitem[Ramesh et~al.(2022)Ramesh, Dhariwal, Nichol, Chu, and
  Chen]{ramesh2022hierarchical}
Aditya Ramesh, Prafulla Dhariwal, Alex Nichol, Casey Chu, and Mark Chen.
\newblock Hierarchical text-conditional image generation with {CLIP} latents.
\newblock \emph{arXiv preprint arXiv:2204.06125}, 1\penalty0 (2):\penalty0 3,
  2022.

\bibitem[Rombach et~al.(2022{\natexlab{a}})Rombach, Blattmann, Lorenz, Esser,
  and Ommer]{Rombach_2022_CVPR}
Robin Rombach, Andreas Blattmann, Dominik Lorenz, Patrick Esser, and Bj\"orn
  Ommer.
\newblock High-resolution image synthesis with latent diffusion models.
\newblock In \emph{Proceedings of the IEEE/CVF Conference on Computer Vision
  and Pattern Recognition (CVPR)}, pages 10684--10695, 2022{\natexlab{a}}.

\bibitem[Rombach et~al.(2022{\natexlab{b}})Rombach, Blattmann, Lorenz, Esser,
  and Ommer]{rombach2022high}
Robin Rombach, Andreas Blattmann, Dominik Lorenz, Patrick Esser, and Bj{\"o}rn
  Ommer.
\newblock High-resolution image synthesis with latent diffusion models.
\newblock In \emph{Proceedings of the IEEE/CVF conference on computer vision
  and pattern recognition}, pages 10684--10695, 2022{\natexlab{b}}.

\bibitem[Saharia et~al.(2022)Saharia, Chan, Saxena, Li, Whang, Denton,
  Ghasemipour, Gontijo~Lopes, Karagol~Ayan, Salimans,
  et~al.]{saharia2022photorealistic}
Chitwan Saharia, William Chan, Saurabh Saxena, Lala Li, Jay Whang, Emily~L
  Denton, Kamyar Ghasemipour, Raphael Gontijo~Lopes, Burcu Karagol~Ayan, Tim
  Salimans, et~al.
\newblock Photorealistic text-to-image diffusion models with deep language
  understanding.
\newblock \emph{Advances in Neural Information Processing Systems},
  35:\penalty0 36479--36494, 2022.

\bibitem[Samuel et~al.(2023{\natexlab{a}})Samuel, Ben-Ari, Darshan, Maron, and
  Chechik]{samuel2023norm}
Dvir Samuel, Rami Ben-Ari, Nir Darshan, Haggai Maron, and Gal Chechik.
\newblock Norm-guided latent space exploration for text-to-image generation.
\newblock \emph{arXiv preprint arXiv:2306.08687}, 2023{\natexlab{a}}.

\bibitem[Samuel et~al.(2023{\natexlab{b}})Samuel, Ben-Ari, Raviv, Darshan, and
  Chechik]{samuel2023all}
Dvir Samuel, Rami Ben-Ari, Simon Raviv, Nir Darshan, and Gal Chechik.
\newblock It is all about where you start: Text-to-image generation with seed
  selection.
\newblock \emph{arXiv preprint arXiv:2304.14530}, 2023{\natexlab{b}}.

\bibitem[Sohl-Dickstein et~al.(2015)Sohl-Dickstein, Weiss, Maheswaranathan, and
  Ganguli]{sohl2015deep}
Jascha Sohl-Dickstein, Eric Weiss, Niru Maheswaranathan, and Surya Ganguli.
\newblock Deep unsupervised learning using nonequilibrium thermodynamics.
\newblock In \emph{International conference on machine learning}, pages
  2256--2265. PMLR, 2015.

\bibitem[Song et~al.(2020)Song, Meng, and Ermon]{song2020denoising}
Jiaming Song, Chenlin Meng, and Stefano Ermon.
\newblock Denoising diffusion implicit models.
\newblock \emph{arXiv preprint arXiv:2010.02502}, 2020.

\bibitem[Song et~al.(2018)Song, Yang, Lin, Liu, Huang, Li, and
  Kuo]{song2018contextual}
Yuhang Song, Chao Yang, Zhe Lin, Xiaofeng Liu, Qin Huang, Hao Li, and C-C~Jay
  Kuo.
\newblock Contextual-based image inpainting: Infer, match, and translate.
\newblock In \emph{Proceedings of the European conference on computer vision
  (ECCV)}, pages 3--19, 2018.

\bibitem[Tumanyan et~al.(2022)Tumanyan, Bar-Tal, Bagon, and
  Dekel]{tumanyan2022splicing}
Narek Tumanyan, Omer Bar-Tal, Shai Bagon, and Tali Dekel.
\newblock Splicing vit features for semantic appearance transfer.
\newblock In \emph{Proceedings of the IEEE/CVF Conference on Computer Vision
  and Pattern Recognition}, pages 10748--10757, 2022.

\bibitem[Tumanyan et~al.(2023)Tumanyan, Geyer, Bagon, and
  Dekel]{Tumanyan_2023_CVPR}
Narek Tumanyan, Michal Geyer, Shai Bagon, and Tali Dekel.
\newblock Plug-and-play diffusion features for text-driven image-to-image
  translation.
\newblock In \emph{Proceedings of the IEEE/CVF Conference on Computer Vision
  and Pattern Recognition (CVPR)}, pages 1921--1930, 2023.

\bibitem[Wang et~al.(2004)Wang, Bovik, Sheikh, and Simoncelli]{wang2004image}
Zhou Wang, Alan~C Bovik, Hamid~R Sheikh, and Eero~P Simoncelli.
\newblock Image quality assessment: from error visibility to structural
  similarity.
\newblock \emph{IEEE transactions on image processing}, 13\penalty0
  (4):\penalty0 600--612, 2004.

\bibitem[Wang et~al.(2022)Wang, Zheng, He, Chen, and Zhou]{wang2022diffusion}
Zhendong Wang, Huangjie Zheng, Pengcheng He, Weizhu Chen, and Mingyuan Zhou.
\newblock Diffusion-{GAN}: Training {GANs} with diffusion.
\newblock \emph{arXiv preprint arXiv:2206.02262}, 2022.

\bibitem[Xu et~al.(2023)Xu, Ma, Huang, Lee, and Chai]{xu2023cyclenet}
Sihan Xu, Ziqiao Ma, Yidong Huang, Honglak Lee, and Joyce Chai.
\newblock {CycleNet}: Rethinking cycle consistent in text‑guided diffusion
  for image manipulation.
\newblock In \emph{Advances in Neural Information Processing Systems
  (NeurIPS)}, 2023.

\bibitem[Yu et~al.(2020)Yu, Chen, Wang, Xian, Chen, Liu, Madhavan, and
  Darrell]{yu2020bdd100k}
Fisher Yu, Haofeng Chen, Xin Wang, Wenqi Xian, Yingying Chen, Fangchen Liu,
  Vashisht Madhavan, and Trevor Darrell.
\newblock Bdd100k: A diverse driving dataset for heterogeneous multitask
  learning.
\newblock In \emph{Proceedings of the IEEE/CVF conference on computer vision
  and pattern recognition}, pages 2636--2645, 2020.

\bibitem[Zhang and Agrawala(2023)]{zhang2023adding}
Lvmin Zhang and Maneesh Agrawala.
\newblock Adding conditional control to text-to-image diffusion models.
\newblock \emph{arXiv preprint arXiv:2302.05543}, 2023.

\bibitem[Zhao et~al.(2020)Zhao, Mo, Lin, Wang, Zuo, Chen, Xing, and
  Lu]{zhao2020uctgan}
Lei Zhao, Qihang Mo, Sihuan Lin, Zhizhong Wang, Zhiwen Zuo, Haibo Chen, Wei
  Xing, and Dongming Lu.
\newblock {UCTGAN}: Diverse image inpainting based on unsupervised cross-space
  translation.
\newblock In \emph{Proceedings of the IEEE/CVF conference on computer vision
  and pattern recognition}, pages 5741--5750, 2020.

\bibitem[Zhu et~al.(2017)Zhu, Park, Isola, and Efros]{zhu2017unpaired}
Jun-Yan Zhu, Taesung Park, Phillip Isola, and Alexei~A Efros.
\newblock Unpaired image-to-image translation using cycle-consistent
  adversarial networks.
\newblock In \emph{Proceedings of the IEEE international conference on computer
  vision}, pages 2223--2232, 2017.

\end{thebibliography}
}

% WARNING: do not forget to delete the supplementary pages from your submission 
% \input{sec/X_suppl}
\appendix
\clearpage
%\setcounter{page}{1}
%\maketitlesupplementary

\section{Implementation details}
\subsection{Fine-tuning Stable Diffusion}
All of the results presented in the main paper were achieved using Stable Diffusion 2.1~\cite{Rombach_2022_CVPR}, specifically fine-tuned for automotive images. The training dataset comprised a random selection of $25k$ day images and $25k$ night images sourced from the BDD100k dataset \cite{yu2020bdd100k}, center cropped to $512\times512$ pixels. Further statistics regarding the internal distribution of other attributes within the sampled data are outlined in \Cref{tab:inner_dist}.

\begin{table}[h!]
    \centering
    \begin{tabular}{c|c|c|}
                       & \textbf{Day} & \textbf{Night}\\
       \hline
       \textbf{Clear}  & 12247 & 20861\\
       \textbf{Rainy}  & 2476 & 2016 \\
       \textbf{Foggy}  & 46 & 57\\
       \textbf{Other}  & 10231 & 2066\\
       \textbf{Undefined}  &  0& 0\\

       \hline
                       & \color{purple}{25k}  & \color{purple}{25k} \\

    \end{tabular}
    \caption{Distribution of attributes in the training dataset.}
    \label{tab:inner_dist}
\end{table}

We used the fine-tuning scheme suggested in ControlNet \cite{zhang2023adding}, with pre-produced Canny maps as the spatial condition. We automatically generated the textual conditions using information provided in BDD100k's metadata logs, regarding the image's Scene, Weather and Time-Of-Day. The resulting prompts have the following form:
\begin{quote}
``A \textcolor{magenta}{Scene} in a \textcolor{magenta}{Weather} \textcolor{magenta}{Time-Of-Day}'' 
\end{quote}

\begin{figure}[h!]
    \centering
    \includegraphics[width=1.\linewidth]{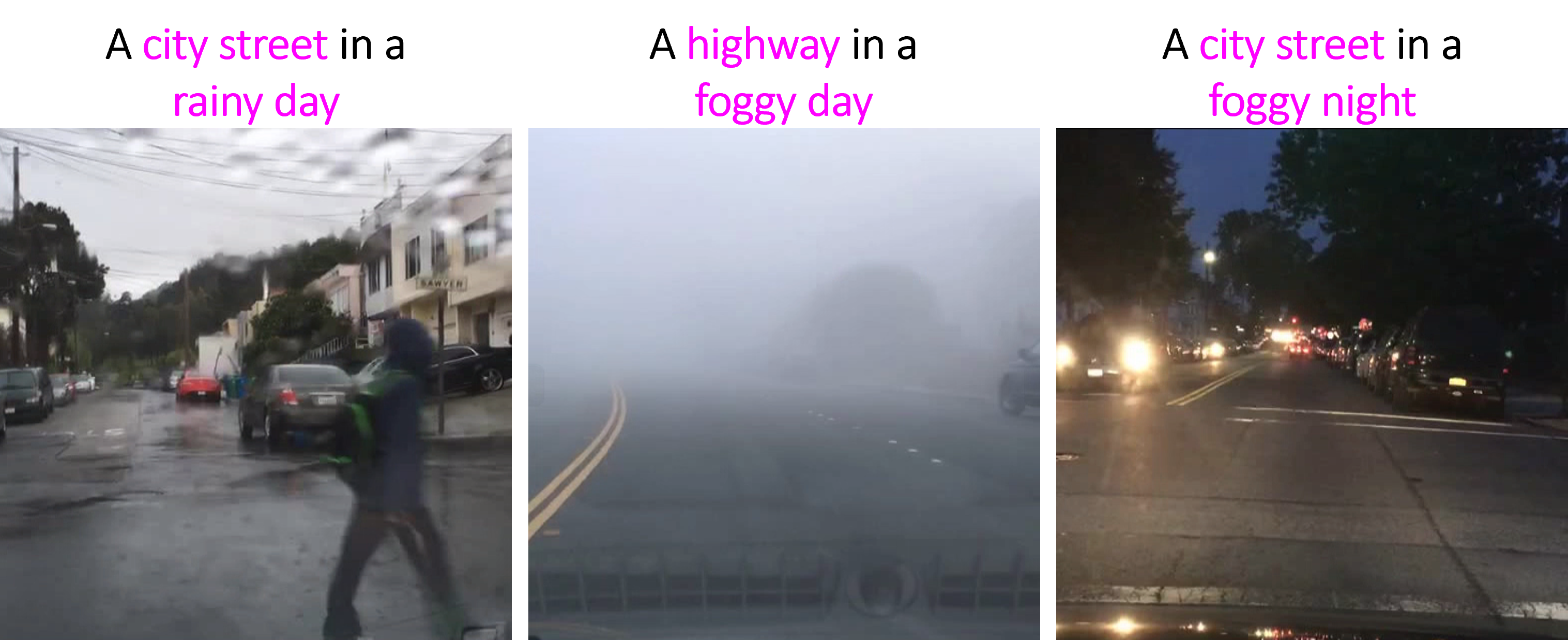}
    \caption{Prompts automatically generated for BDD100k images using the attached metadata.}
    \label{fig:prompt}
\end{figure}

The various choices available in the metadata logs of BDD100k for individual attributes are delineated in \Cref{tab:attr}. It should be noted that all images featuring an ``undefined'' label for any attribute have been excluded from the training set. \Cref{fig:prompt} illustrates a few automated textual prompts and the corresponding images, as examples.

\begin{table}[h!]
    \centering
    \begin{tabular}{c|p{5cm}}
         \hline
         \textbf{Scene}&  tunnel, residential, parking lot, undefined, city street, gas stations, highway \\
         \hline
         \textbf{Weather}&  
         rainy, snowy, clear, overcast, undefined, partly cloudy, foggy\\
         \hline
         \textbf{Time-Of-Day}&  daytime, night, dawn/dusk, undefined\\
         \hline
    \end{tabular}
    \caption{Attributes and corresponding options provided in BDD100k metadata logs.}
    \label{tab:attr}
\end{table}

\begin{figure*}[h!]
    \centering
    \includegraphics[width=0.6\linewidth]{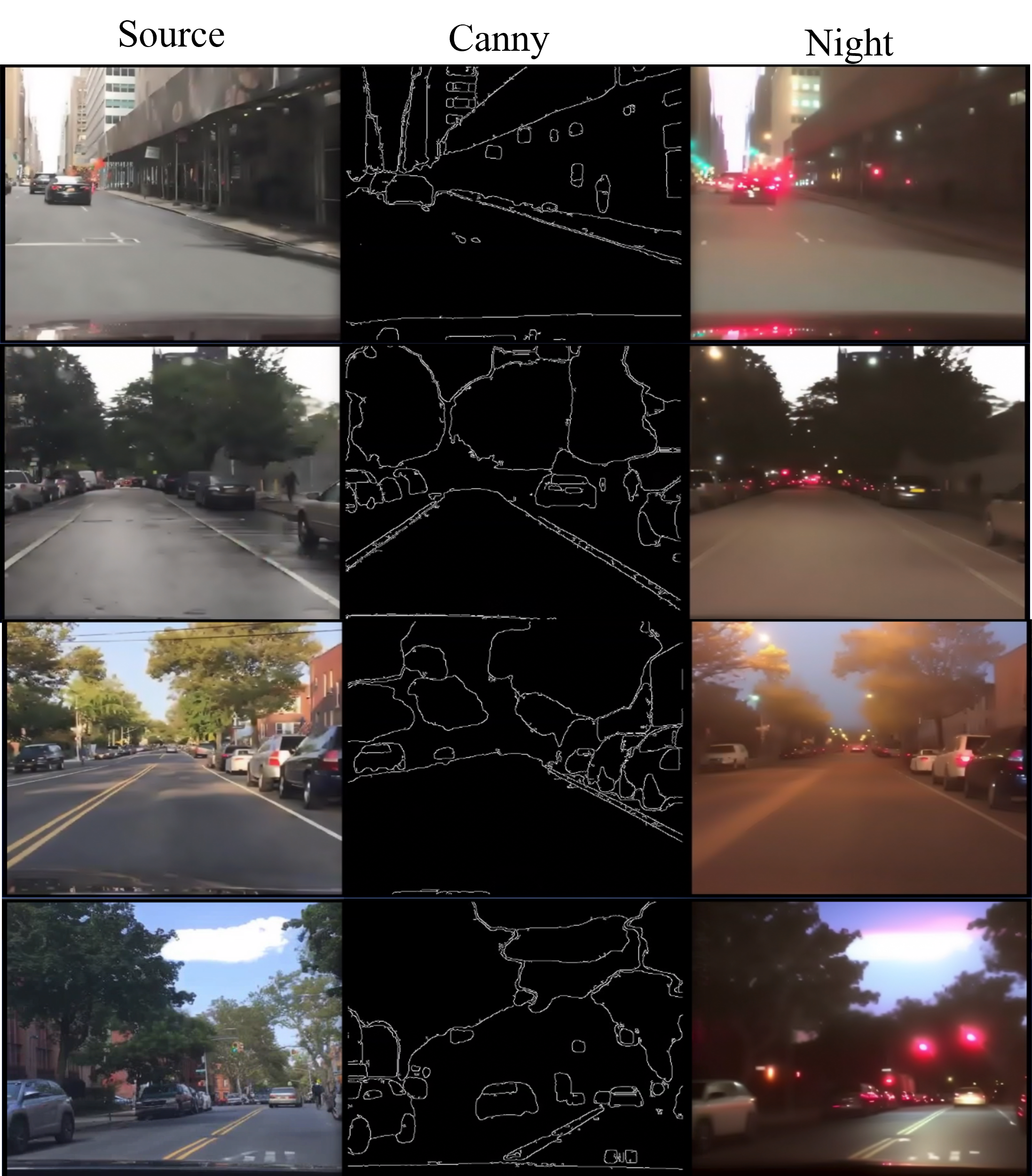}
    \caption{Naive ablation: using a fine-tuned SD 2.1 without Seed Translation and Trajectory Optimization.}
    \label{fig:abl}
\end{figure*}

It is essential to highlight that, during the fine-tuning phase, our network was trained solely to reconstruct the training images, without engaging in any form of translation. The ControlNet \cite{zhang2023adding} fine-tuning mechanism, was specifically employed for adapting to Automotive (BDD100k) domain rather than for the purposes of image editing or translation. Figure \ref{fig:abl} showcases the fine-tuned model's ability to transform daytime images into a nighttime appearance without employing the suggested ST and TO blocks. Specifically, we utilized the fine-tuned SD 2.1 model to invert a source daytime image into a seed and then utilized the model to generate an output from that seed using the Canny map of the source image and a target-domain-related prompt. It can be observed that the outcomes suffer from inaccuracies in terms of both content preservation and achieving the desired appearance in the target domain. This observation confirms that the translation capabilities of our approach, demonstrated in the main paper and in \Cref{fig:dm_comp,fig:gan_comp} below, were achieved due to the use of seed optimization and trajectory translation.

\subsection{User Study Details}
\begin{figure*}[h!]
    \centering
    \includegraphics[width=0.7\linewidth]{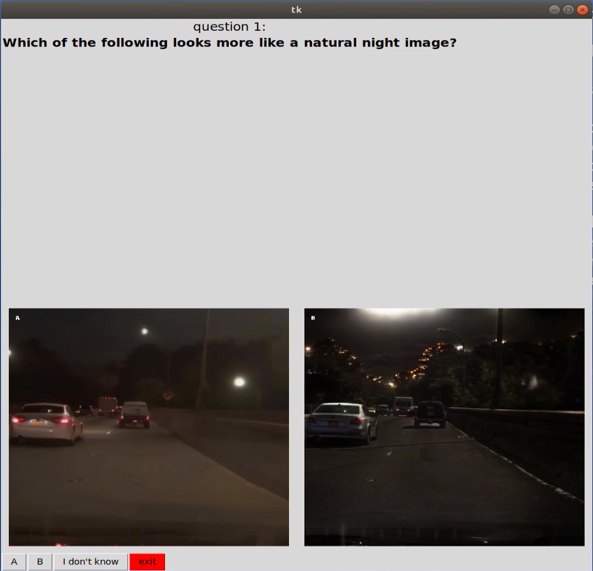}
    \caption{User Study UI screenshot, Q1.}
    \label{fig:ill_mos1}
\end{figure*}

\begin{figure*}[h!]
    \centering
    \includegraphics[width=0.7\linewidth]{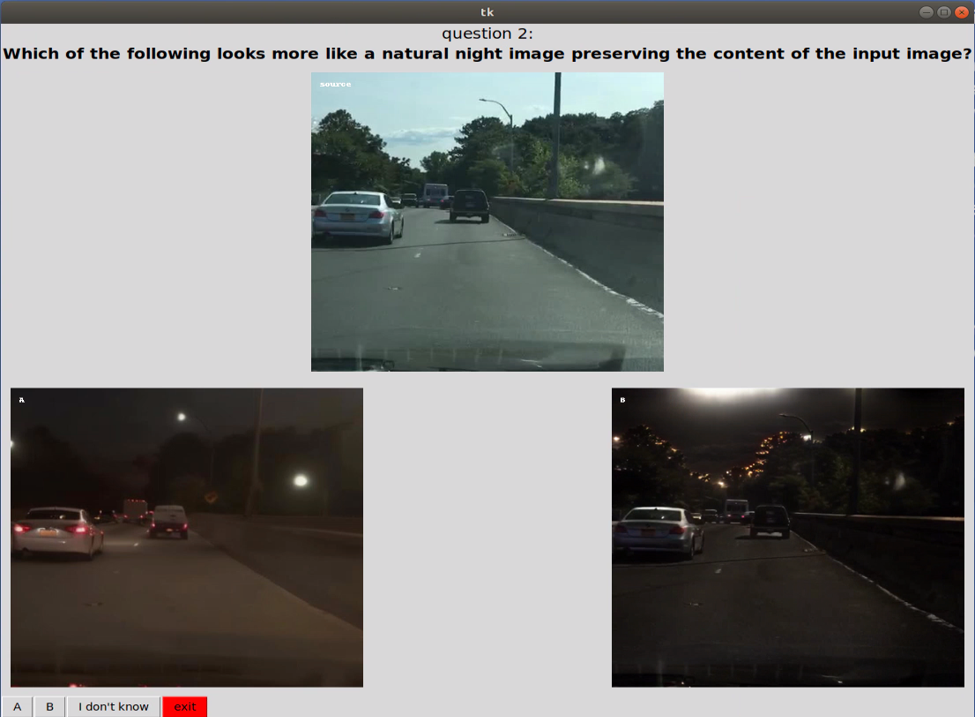}
    \caption{User Study UI screenshot, Q2.}
    \label{fig:ill_mos2}
\end{figure*}

We performed our user study based on the mean opinion of 14 independent participants, between the ages of 30 to 60, all have a normal or corrected to normal vision. In total, we collected $\sim1300$ answers, with the results detailed in Figure 4 in the main text. Screenshots from our user study (for Q1 and Q2) are provided in \Cref{fig:ill_mos1,fig:ill_mos2}, respectively.

\section{Additional Qualitative Results}

Here, we present a number of additional qualitative results, ablations, and comparisons to various baselines.

\begin{figure*}[h!]
    \centering
    \includegraphics[width=0.52\linewidth]{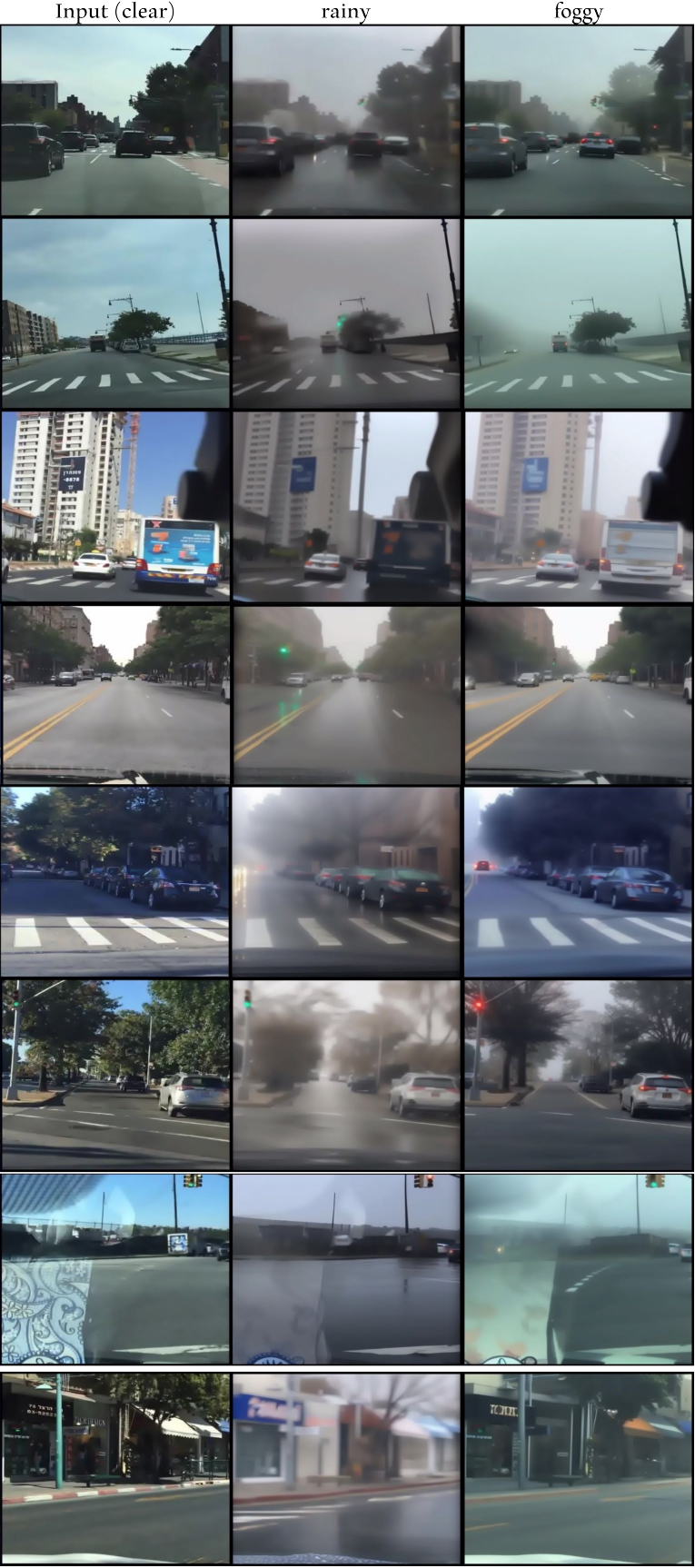}
    \caption{Weather translation using S2ST, where a source clear image is being translated into rainy/foggy.}
    \label{fig:rain-fog}
\end{figure*}

\Cref{fig:rain-fog} presents several examples of clear-to-rainy and clear-to-foggy translations, using S2ST. It is important to highlight the limited availability of rainy and foggy images within the domain adaptation (fine-tuning) dataset, as evidenced by Table \ref{tab:inner_dist}, in contrast to the abundance of clear images. This disparity in sample distribution is attributed to the scarcity of samples in the BDD100k dataset. Yet, our method is able to generate plausible translations to these domains as well. 

\begin{figure*}[h!]
    \centering
    \includegraphics[width=0.52\linewidth]{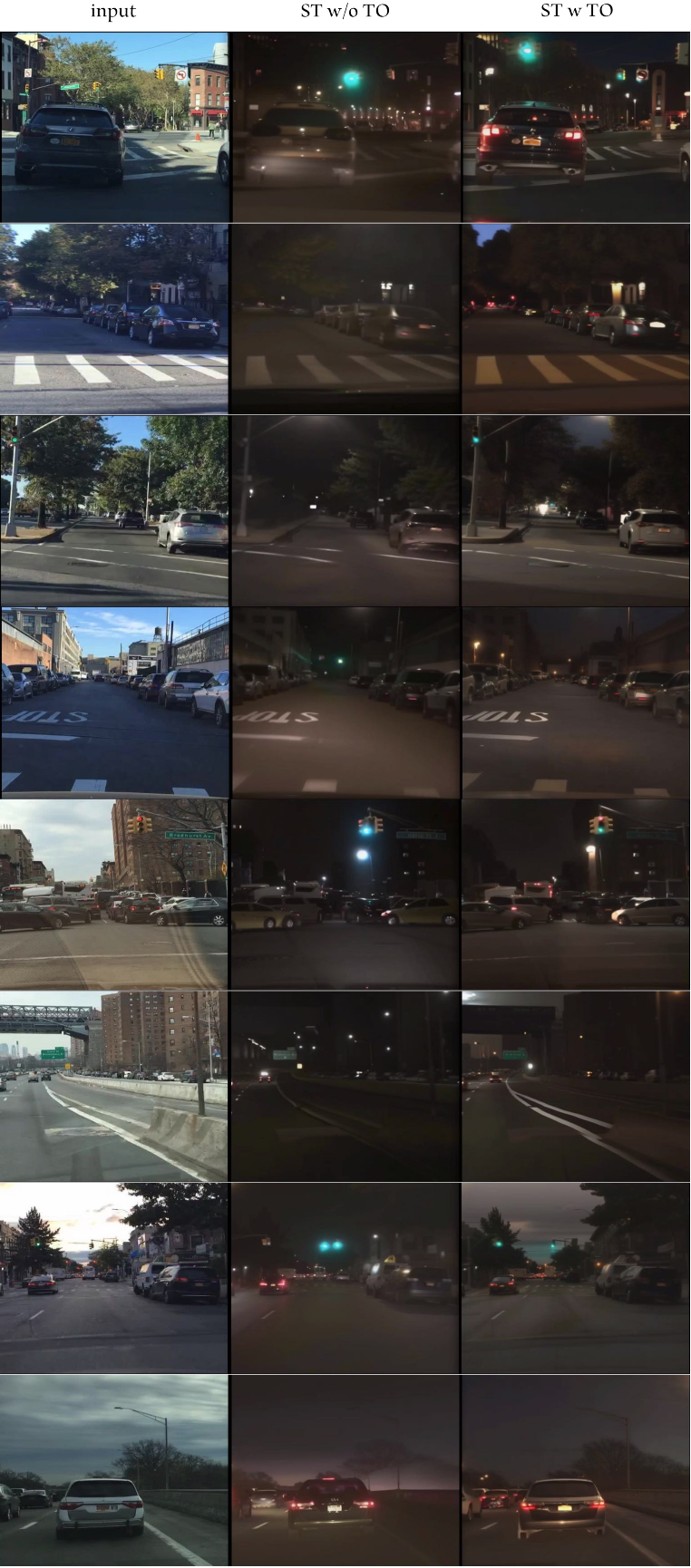}
    \caption{More examples of the effect of the two components of our model: Seed Translation (ST, middle column) and Trajectory Optimization (TO, right column). While the main domain shift is performed during the ST phase, the TO phase enhances the image and the preservation of details with respect to the input image.}
    \label{fig:ablation2}
\end{figure*}

\Cref{fig:ablation2} presents additional examples of the effect of the two components of our model: Seed Translation (ST, middle column) and Trajectory Optimization (TO, right column). While the main domain shift is performed during the ST phase, the TO phase enhances the image and the preservation of details with respect to the input image.

\begin{figure*}[h!]
    \centering
    \includegraphics[height=0.98\textheight]{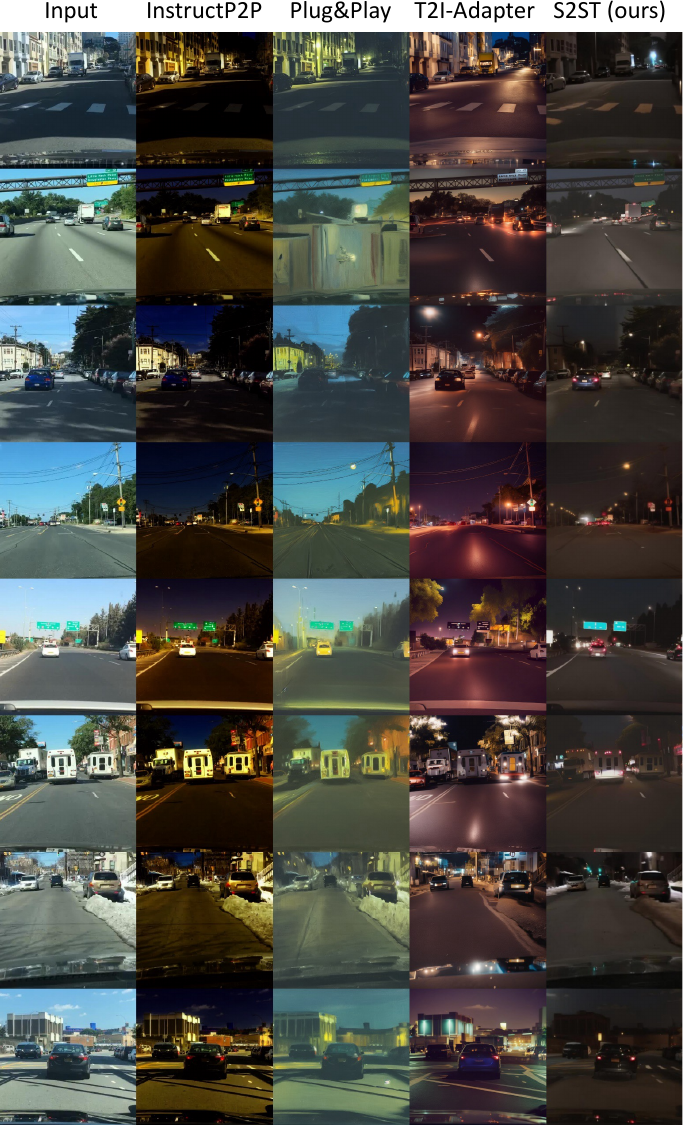}
    \caption{Qualitative Comparison to DM based Methods}
    \label{fig:dm_comp}
\end{figure*}

\Cref{fig:dm_comp} present qualitative comparisons to several DM-based methods: Instruct-Pix2Pix~\cite{brooks2023instructpix2pix}, Plug-and-Play diffusion features~\cite{Tumanyan_2023_CVPR}, and T2I adapter~\cite{mou2023t2iadapter}. All of these methods are based on Stable Diffusion~\cite{Rombach_2022_CVPR}, and are capable of various image-to-image translation and image editing tasks, but neither of them aims specifically at day-to-night translations or automotive scenes. It may be seen that Instruct-Pix2pix and Plug-and-Play tend to retain the sharp shadows cast by the sun on the road surface, and some other lighting effects from the input daytime images. In addition, all methods result in less realistic night images, compared to S2ST. 

\begin{figure*}[h!]
    \centering
    \includegraphics[height=0.98\textheight]{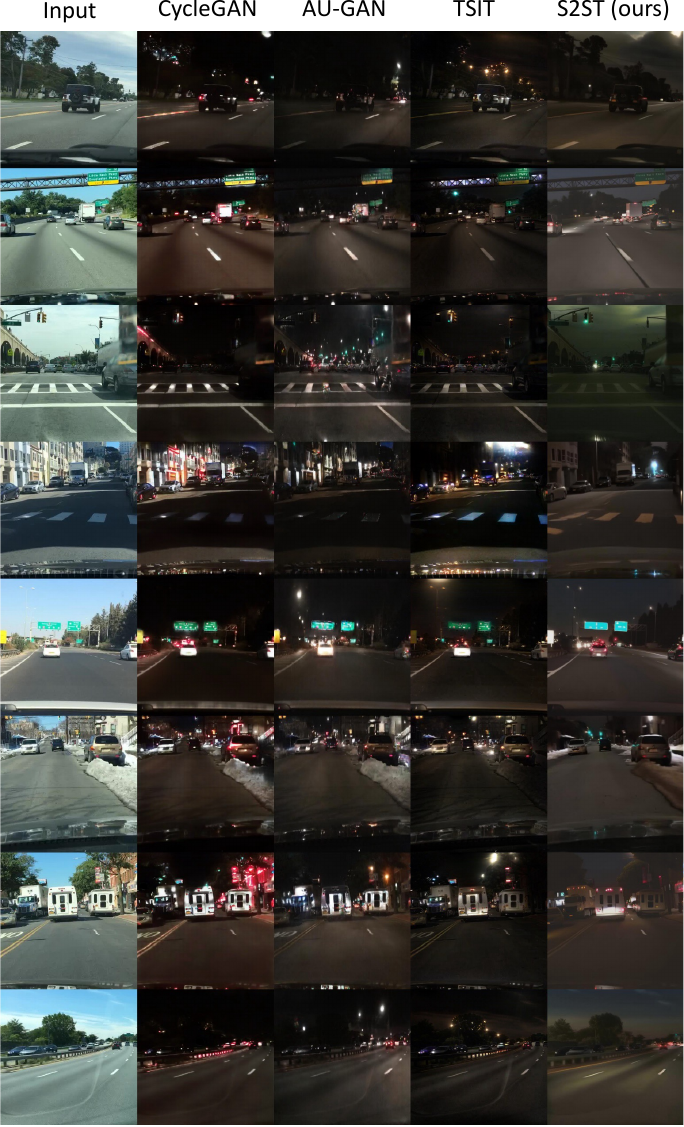}
    \caption{Qualitative Comparison to GAN based Methods.}
    \label{fig:gan_comp}
\end{figure*}

\Cref{fig:gan_comp} presents additional qualitative comparisons to the GAN-based approaches of CycleGAN~\cite{zhu2017unpaired}, AU-GAN~\cite{kwak2021adverse}, and TSIT~\cite{jiang2020tsit}. As discussed in the main paper, these methods rely on a discriminator, and thus favor presence of local features that are characteristic of nighttime scenes at the expense of enforcing global semantic consistency, such as correct placement of light sources.

\clearpage

%\clearpage
%\section{Additional Results}
%\Or{in the following pages we will present figures with more results and comparisons. I think that no text is needed but the captions. For now, I left a title in every page just to specify what figure should be placed in each page. I think that we can omit the titles once we insert the figures.}

\clearpage

\clearpage

\clearpage

\clearpage

\clearpage

\end{document}